\documentclass[10pt,twocolumn,letterpaper]{article}

\PassOptionsToPackage{table}{xcolor}

\usepackage[pagenumbers]{cvpr} % To force page numbers, e.g. for an arXiv version

\usepackage{multirow}
\usepackage{multicol}
\usepackage{amsmath}
\usepackage{amssymb}
\usepackage{makecell}
\usepackage{tabu, booktabs}
\usepackage{subcaption}
\usepackage[table]{xcolor}
\usepackage[utf8]{inputenc}
\usepackage{siunitx}

\usepackage{colortbl}
\usepackage{array}
\usepackage{tabularx}
\usepackage{tikz}
\usepackage{url}

\usepackage{amsthm}

\usepackage[accsupp]{axessibility}  % Improves PDF readability for those with disabilities.

\usepackage[many]{tcolorbox}
\tcbuselibrary{skins}

\usepackage{listings}

\definecolor{cvprblue}{rgb}{0.21,0.49,0.74}
\usepackage[pagebackref,breaklinks,colorlinks,allcolors=cvprblue]{hyperref}

\def\paperID{***} % *** Enter the Paper ID here

\def\confName{CVPR}
\def\confYear{2026}

\title{Reinforcing Structured Chain-of-Thought for Video Understanding}

\author{
Peiyao Wang$^{1}$\thanks{This work was conducted during an internship at Amazon. Email:~peiyaowang@cs.stonybrook.edu.} \quad
Haotian Xu$^{2}$ \quad
Noranart Vesdapunt$^{2}$ \quad
Rui Hou$^{2}$ \quad
Jingyi Zhang$^{2}$ \\
Haibin Ling$^{1}$\thanks{Corresponding authors: Haibin Ling (hling@cs.stonybrook.edu) and Kah Kuen Fu (kahkuen@amazon.com). Haibin Ling was involved in this work while affiliated with Stony Brook University.} \quad
Oleksandr Obiednikov$^{2}$ \quad
Ning Zhou$^{2}$ \quad
Kah Kuen Fu$^{2}$\footnotemark[2]\\[0.4em]
$^{1}$Stony Brook University \quad
$^{2}$Amazon\\[0.25em]
}

\begin{document}
\maketitle
\begin{abstract}
Multi-modal Large Language Models (MLLMs) show promise in video understanding. However, their reasoning often suffers from thinking drift and weak temporal comprehension, even when enhanced by Reinforcement Learning (RL) techniques like Group Relative Policy Optimization (GRPO). Moreover, existing RL methods usually depend on Supervised Fine-Tuning (SFT), which requires costly Chain-of-Thought (CoT) annotation and multi-stage training, and enforces fixed reasoning paths, limiting MLLMs' ability to generalize and potentially inducing bias.
To overcome these limitations, we introduce \textbf{S}ummary-\textbf{D}riven \textbf{R}einforcement \textbf{L}earning (\textbf{SDRL}), a novel single-stage RL framework that obviates the need for SFT by utilizing a Structured CoT format: Summarize $\rightarrow$ Think $\rightarrow$ Answer.
SDRL introduces two self-supervised mechanisms integrated into the GRPO objective: 1) Consistency of Vision Knowledge (CVK) enforces factual grounding by reducing KL divergence among generated summaries; and 2) Dynamic Variety of Reasoning (DVR) promotes exploration by dynamically modulating thinking diversity based on group accuracy. This novel integration effectively balances alignment and exploration, supervising both the final answer and the reasoning process.
Our method achieves state-of-the-art performance on seven public VideoQA datasets.
% Additionally, we construct and will release an 80K VideoQA dataset focusing action sequencing and temporal causality. 

% \textcolor{red}{We release the code and dataset at \url{https://github.com/xxx/SDRL}.}

%Experiments on MMVU, TempCompass, and VideoMME show promising accuracy gains.
%confirming the effectiveness of our structured reasoning supervision. 

% Specifically, our method first analyzes the characteristics of high-quality CoT for video understanding and then introduces a structured reasoning format: Summarize, Think, Answer, to guide the reasoning trajectory. The crucial component is the Summary stage, which explicitly enforces the temporal action order. To optimize this, we introduce group consistency across multiple sampled summaries while encouraging dynamic diversity among reasoning paths during the Think stage. This strategy effectively balances alignment (consistency) and exploration (diversity) within the RL loop, supervising both the correctness of the final answer and the fidelity of the reasoning process.
% and group-based RL effectively enhance the reliability and interpretability of video-language models.

%\lingg{Maybe we can call our method SuTA (Summary-Think-Anaswer)?}
\end{abstract}    
\section{Introduction}
\label{sec:intro}

\begin{figure}
\includegraphics[width=1.0\linewidth]{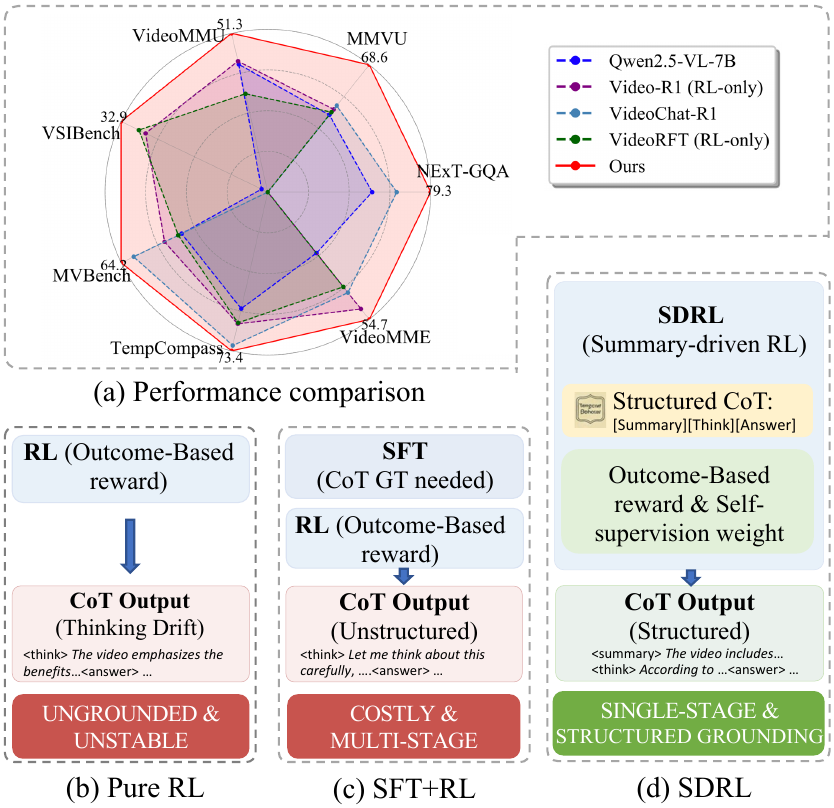}
    \caption{Performance comparison and training paradigms for video reasoning models.
(a) Performance comparison among several benchmarks.
(b)-(d) Training paradigm analysis: (b) Pure RL often yields ungrounded and unstable Chain-of-Thought (CoT) outputs. (c) SFT+RL is costly and complex. In contrast, (d) SDRL (Summary-driven RL) utilizes a Structured CoT and self-supervision to achieve stable and grounded video reasoning.}
    \label{fig:intro}
\end{figure}

% peiyao's version 3
Multimodal Large Language Models (MLLMs) have significantly advanced the frontier of video understanding, enabling open-ended reasoning over dynamic visual scenes~\cite{feng2025videor1, li2025videochatr1, dang2025twgrpo, yu2025videoutr, bai2025qwen25vl}. 
% By jointly modeling visual and textual modalities, these models establish a holistic paradigm over traditional modular systems, achieving impressive results in tasks such as Video Question Answering (VideoQA), temporal grounding, and captioning. 
The capability of MLLMs is further amplified by the Chain-of-Thought (CoT) prompting technique~\cite{shao2024visual, wang2025multimodal}. By explicitly introducing intermediate reasoning steps, CoT enhances both interpretability and logical reasoning, allowing models to ``think before answering." However, realizing the full potential of CoT often requires high-quality reasoning data for effective training~\cite{wei2022chain,zhang2024chain,han2025videoespresso}.

A significant recent advancement involves leveraging Reinforcement Learning (RL), such as Group Relative Policy Optimization (GRPO)~\cite{shao2024deepseekmath}, to enhance MLLMs' complex reasoning abilities~\cite{zhang2025right}. By optimizing models with reward signals based on verifiable outcomes (\textit{e.g.}, final answer correctness), RL offers a scalable path to elicit beneficial problem-solving strategies without the need for extensive CoT labels. However, this outcome-driven solution is fundamentally limited in complex video tasks: (1) \textit{Thinking drift from unconstrained reasoning}: Relying solely on the final reward leaves intermediate reasoning steps unconstrained. This often leads to thinking drift~\cite{luo2025videover}, where the model generates verbose or reasoning irrelevant with the visual evidence, significantly hindering result stability.  (2) \textit{Weak temporal reasoning}: %This issue is compounded by the underlying visual architecture. 
MLLMs frequently represent video as stacked or averaged frame embeddings, hence ignoring fine-grained temporal dependencies. As recent studies demonstrate~\cite{feng2025breaking}, such lack of temporal awareness causes significantly poor performance on temporally-sensitive VideoQA tasks.

An alternative to direct RL is to add imitation learning, often implemented via Supervised Fine-Tuning (SFT) on expert demonstrations~\cite{chen2025sft,li2025vla,zhai2024fine}. SFT is used to instill targeted reasoning behaviors, compensating for the random exploration phase in RL. For instance, \cite{ni2025point, wu2025grounded} use SFT to inject explicit spatio-temporal information or integrate descriptive captions into CoT to enhance grounding. While SFT can distill valuable reasoning behaviors, its next-token prediction objective enforces rigid, token-level imitation, limiting the model’s generalization beyond training data. Consequently, long and complex demonstrations often lead to overfitting and shallow reasoning. Moreover, most approaches with such SFT+RL pipeline require costly annotations, are time-consuming, and may potentially constrain the base model's intrinsic reasoning potential~\cite{liu2025uft}. 

% Recent works \ling{just with RL constraining reasoning focusing} (e.g., TW-GRPO) show comparable results to SFT $\rightarrow$ RL, questioning the necessity of SFT before RL for structured reasoning in video understanding. %raising a critical question: Is SFT truly necessary before RL for structured reasoning in video understanding?

%\lingg{Add necessary citations in the above paragraphs.}

This critical research gap, the need for efficient, structurally supervised, and temporally grounded training, motivates our work. We introduce Summary-Driven Reinforcement Learning (SDRL), a novel single-stage framework designed to enhance the temporal action order fidelity and interpretability of MLLMs without the need for prior SFT. Our core innovation lies in the direct integration of a Structured CoT into the RL objective. Specifically, we propose a Summarize$\rightarrow$Think$\rightarrow$Answer structure. The Summarize stage explicitly mandates the correct temporal action order, serving as a robust, structure-based anchor that grounds the subsequent reasoning. To effectively optimize this structure with RL, we leverage a novel GRPO-based objective that balances alignment and exploration. This objective enhances group consistency across sampled summaries while encouraging dynamic diversity during the Think stage.

% Our main contributions are summarized as follows:
% \begin{itemize}
% \item
% We propose Summary-Driven Reinforcement Learning (SDRL), a single-stage RL framework that integrates a novel Structured CoT to enhance temporal action order fidelity, eliminating the need for SFT pre-training.
% \item We introduce the Summarize, Think, Answer structure. We develop Summary Consistency Loss (SCL) to enable structural supervision without labeled CoT data and Dynamic Thinking Loss (DTL) to encourage efficient deep thinking, which collectively ground the reasoning and alleviate thinking drift.
% \item We construct a 50K+ VideoQA dataset combining existing benchmarks with newly annotated temporal reasoning tasks. SDRL achieves state-of-the-art results on several public datasets (e.g., MMVU, TempCompass, and VideoMME), while supporting explainable AI as errors can be traced back to the generated factual summary. We will also release our dataset to support this emerging research direction.

Our main contributions are as follows:
\begin{itemize}

\item
We propose Summary-Driven Reinforcement Learning (SDRL), which utilizes a Structured CoT format (Summarize$\rightarrow$Think$\rightarrow$Answer) to enhance temporal and factual reasoning fidelity, effectively recuding the reliance on SFT. % for supervised instruction tuning (SFT). 
% \item We simplify the training recipe from common two-stage SFT+RL to single-stage RL only by introducing two complementary self-supervised structural constraints and integrate them directly into the policy objective to stabilize and diversify reasoning.
% \item We construct a 53K VideoQA dataset combining existing benchmarks with newly annotated temporal reasoning tasks to support structured video reasoning research.
\item We introduce two complementary mechanisms: Consistency of Vision Knowledge (CVK) to enforce factual grounding via group-level summary alignment, and Dynamic Variety of Reasoning (DVR) to promote exploration by modulating reasoning diversity.

\item We simplify a two-stage SFT+RL pipeline to a single-stage RL-only framework, leading to superior performance across seven public VideoQA benchmarks.

\end{itemize}

\section{Related Work}
\label{sec:related}
\begin{figure*}[t]
    \centering
    \includegraphics[width=0.95\linewidth]{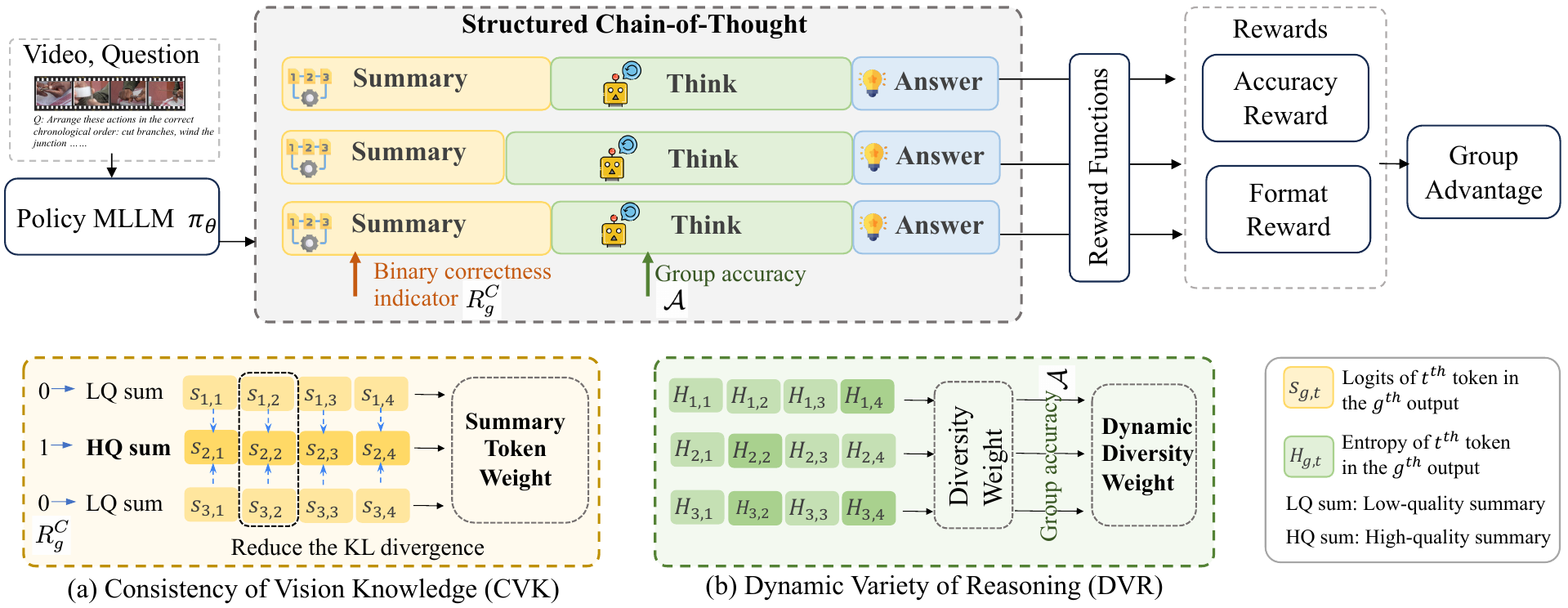}
    \caption{Overview of the SDRL Framework, which introduces Structured Chain-of-Thought. The Policy Model ($\pi_{\theta}$) generates $G$ reasoning sequences, each structured as Summary, Think, and Answer. The framework introduces two structured objectives implemented via token-wise weight: (a) Consistency of Vision Knowledge (CVK) and (b) Dynamic Variety of Reasoning (DVR). These structured weights, along with standard rewards (Accuracy, Format), are combined to derive the group advantage for policy optimization.}
    \label{fig:pip}
\end{figure*}

%Overview of the Structured Chain-of-Thought (SCoT) Optimization Framework. The policy MLLM ($\pi_\theta$) generates $G$ reasoning sequences composed of <summary>, <think>, and <answer>. Two token-level objectives guide learning: (1) Consistency of Vision Knowledge (CVK), which uses binary correctness to identify high-quality summaries and reduces KL divergence across the group, producing KL-penalized summary weights; and (2) Dynamic Variety of Reasoning (DVR), which computes entropy in the thinking stage and scales it by group accuracy to encourage diverse reasoning, yielding entropy-scaled weights. These structured weights, together with standard similarity, accuracy, and format rewards, form the Group Advantage for policy optimization.

\textbf{Reinforcement Learning for MLLMs in Video Understanding.}
% Reinforcement learning (RL) has become a crucial paradigm for enhancing reasoning in large language models (LLMs). Recent systems such as OpenAI-o1~\cite{jaech2024openaio1} and DeepSeek-R1~\cite{guo2025deepseekr1} demonstrate that GRPO-style optimization significantly improves multi-step reasoning, further scaled by works like DeepSeekMath~\cite{shao2024deepseekmath} and Kimi K1.5~\cite{du2025kimi}. Extending this to multimodal settings, RL-based methods~\cite{zhou2025r1zero,liu2025segzero,zhan2025visionr1,deng2025curriculum,peng2025lmmr1,liu2025visualrft,yang2025r1onevision,zhang2025r1vl,deng2025openvlthinker} have incorporated verifiable and rule-based reward mechanisms to improve factual grounding and visual reasoning without human preference data.
%
%Reinforcement Learning has become a crucial technique for enhancing reasoning in MLLMs. Systems like OpenAI-o1~\cite{jaech2024openaio1} and DeepSeek-R1~\cite{guo2025deepseekr1} demonstrate that GRPO-style optimization significantly improves multi-step reasoning, an effect scaled by works such as DeepSeekMath~\cite{shao2024deepseekmath} and Kimi K1.5~\cite{du2025kimi}. %Extending this to the multimodal domain, various RL-based methods have incorporated verifiable and rule-based reward mechanisms to improve factual grounding and visual reasoning without relying on human preference data.
Reinforcement Learning, especially GRPO, has been widely applied to improve LLMs' reasoning capacity~\cite{jaech2024openaio1,guo2025deepseekr1,du2025kimi,shao2024deepseekmath}. Recent works extend GRPO to spatiotemporal reasoning in videos. R1-Omni~\cite{zhao2025r1omni}, Video-R1~\cite{feng2025videor1}, and AoT~\cite{xue2025aot} reveal the benefits of temporal consistency and implicit reasoning rewards, while Video-VER~\cite{luo2025videover} grounds reasoning evidentially. However, they rely on multi-stage SFT+RL pipelines for best performance, which potentially constrain exploration and lead to overfitting to training reasoning patterns.
VideoChat-R1~\cite{li2025videochatr1} introduces an IoU-based soft reward for grounding, TW-GRPO~\cite{dang2025twgrpo} focuses on temporal credit assignment, and GRPO-CARE~\cite{chen2025grpocare} enforces group-level consistency. These efforts highlight the potential of process-aware RL for robust and coherent video reasoning.

\vspace{1mm}\noindent\textbf{Process Supervision and Verification of CoT.}
Recent progress in large language models (LLMs) has shifted from outcome-based optimization toward process-level supervision, which explicitly monitors and evaluates intermediate reasoning steps rather than only final predictions. Early studies~\cite{openai2023process,lightman2024verify} proposed process reward models (PRMs) that assign feedback to each reasoning step, improving interpretability and reasoning fidelity. Subsequent extensions, \textit{e.g.}, PSPO*~\cite{li2024pspo}, LongRePS~\cite{zhu2025longreps} and ThinkPRM~\cite{liu2025thinkprm}, incorporated non-linear reward shaping, long-context reasoning, and automatic verification to scale process supervision without exhaustive human annotation.
While these methods enhance a model’s reasoning capacity, they typically require expensive annotation of intermediate reasoning steps, and they frequently suffer from noisy or mis-aligned process-step rewards~\cite{yuan2024free,zhang2025groundedprm}. In contrast, our proposed method provides self-supervision on the reasoning process without the need for explicit process annotations.

\section{Method}
% Our method, Summary-Driven RL (SDRL), adapts the Group Relative Policy Optimization (GRPO) framework to achieve the Structured Chain-of-Thought (SCoT) framework, enabling robust reasoning without relying on traditional supervised instruction tuning (SFT). SDRL addresses the dual challenges of factual grounding and reasoning diversity in video understanding through a top-down structure ($\text{Summary},\text{Thinking}, \text{Answer}$) and two novel, self-supervised objectives: (1) Consistency of Vision Knowledge (CVK) (Section~\ref{sec:cvk}): aligns generated summaries with factual reality to ensure stability. (2) Dynamic Variety of Reasoning (DVR) (Section~\ref{sec:dvr}): selectively encourages diverse thinking paths to enhance robustness. These constraints are integrated into the Structured Policy Objective (Section~\ref{sec:policy}) by applying token-wise weights to the $\text{Summary}$ and $\text{Thinking}$ segments, modularizing the policy update and effectively steering the model.

Our method, Summary-Driven RL (SDRL), extends GRPO to realize the Structured CoT paradigm, enabling robust reasoning without supervised instruction tuning. Addtionally, SDRL proposes two self-supervised objectives: Consistency of Vision Knowledge (CVK) (Sec.~\ref{sec:cvk}), which aligns summaries with factual content, and Dynamic Variety of Reasoning (DVR) (Sec.~\ref{sec:dvr}), which promotes diverse reasoning paths. These constraints are unified under the Structured Policy Objective (Sec.~\ref{sec:policy}) via token-wise weighting over Summary and Thinking segments.

\subsection{Structured CoT for Top-down Reasoning}

\begin{figure}
    \centering
\includegraphics[width=0.97\linewidth]{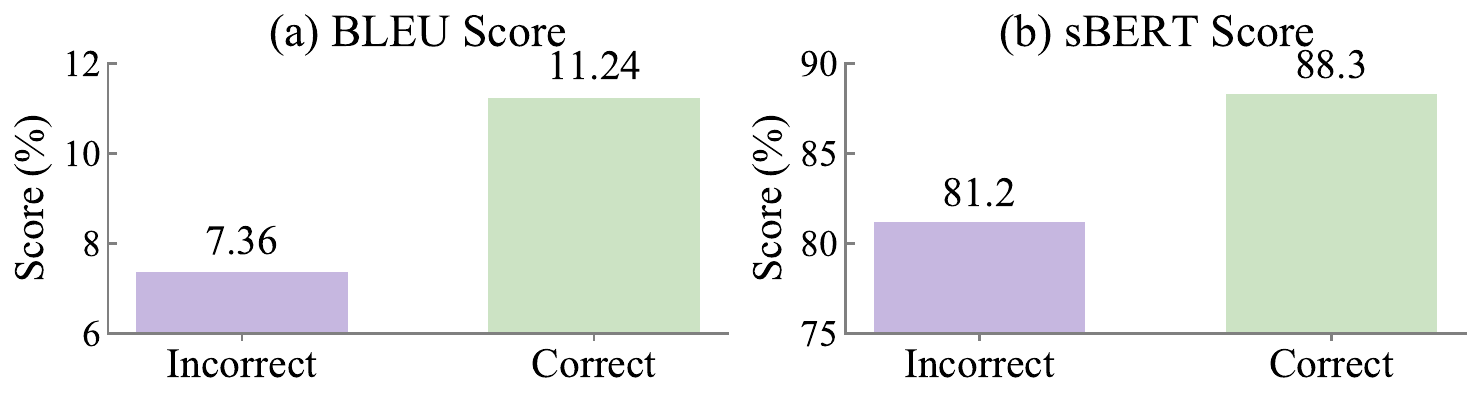}
\vspace{-1mm}
    \caption{BLEU and sBERT scores between different predictions.}
%Correct samples consistently achieve higher scores, indicating stronger alignment with ground-truth semantics.}
    \label{fig:CoT_comp}
\end{figure}

\textbf{What constitutes an effective CoT in video understanding?} 
% The quality of a model’s final prediction is tightly coupled with the fidelity of its intermediate reasoning steps. 
% In video understanding, it is essential that the model's CoT explicitly captures the key actions\cite{?} and their correct temporal order\cite{?}.
% To further verify this dependence, we construct ground-truth CoT sequences enriched with visual and temporal annotations, and then analysis the similarity (measured by BLEU and SBERT scores) between the model-generated CoT and the ground-truth CoT under both correct and incorrect predictions. As illustrated in Figure~\ref{fig:CoT_comp}, correct predictions consistently exhibit a higher similarity between the generated CoT and the ground-truth CoT than incorrect predictions. This analysis reveals a trend: superior performance correlates with CoT sequences that more closely align with the underlying ground-truth reasoning.
% Therefore, we claim that an effective CoT for video understanding must explicitly capture two core components: (1) the key actions or events that occurred in the video, and (2) the temporal order in which these events unfold.
The quality of a model's final prediction is tightly coupled with the fidelity of its intermediate reasoning steps. In video understanding, an effective CoT must explicitly capture the key actions and their correct temporal order.
To verify this dependence, we analyzed the similarity (using BLEU~\cite{papineni2002bleu} and sBERT~\cite{reimers2019sentence} scores) between the model-generated CoT and ground-truth CoT\footnote{The ground-truth CoT sequences were constructed by annotating key actions and their correct temporal order.} under both correct and incorrect final predictions. As shown in Fig. \ref{fig:CoT_comp}, correct predictions consistently exhibit higher CoT-to-ground-truth similarity than incorrect predictions. This analysis reveals a trend: superior performance correlates with CoT sequences that more closely align with the underlying factual reasoning.
Therefore, we argue that an effective CoT for video understanding can explicitly capture two core components: \textbf{(1) the key actions or events}, and \textbf{(2) the temporal order} in which these events unfold. Motivated by this, we introduce a Structured CoT format enforcing a top-down process: \textbf{Summary} (salient, ordered events) $\rightarrow$ logical \textbf{Thinking} $\rightarrow$ final \textbf{Answer}.

\begin{figure}[t]
    \centering
    \includegraphics[width=0.96\linewidth]{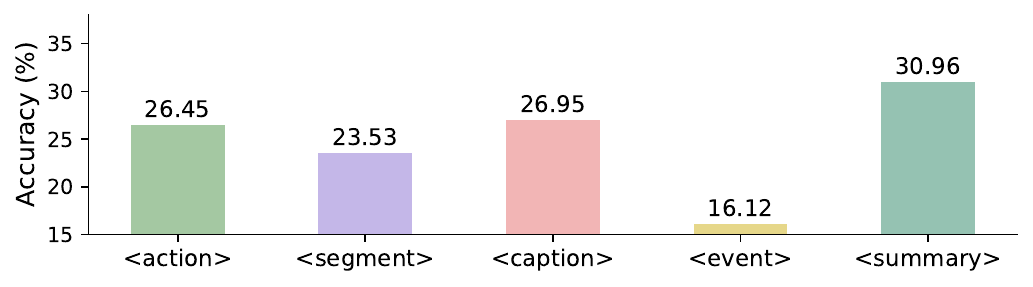}
    \caption{Accuracy comparison across different tag types under training-free, inference-only settings for selecting an appropriate structural format prior to RL optimization.}
%     \caption{Accuracy comparison across different tag types.
% The \textless summary\textgreater~tag most effectively activates the model’s inherent reasoning capability, leading to more coherent and informative CoT generations compared to other tags.}
    \label{fig:tag}
\end{figure}

\vspace{1mm}\noindent\textbf{How should the summary be obtained?}
The procedure for generating the summary depends on annotation availability.
(1) With Ground-Truth Annotations:
When fine-grained temporal labels (e.g., action segments with boundaries) are available, the summary is deterministically constructed from the ground truth.
(2) Without Ground-Truth (Self-Exploration):
When such annotations are unavailable, the model leverages its intrinsic ability to extract high-level temporal cues through self-exploration. To facilitate this, according to the empirical results in Figure~\ref{fig:tag}, we prepend a dedicated \textless summary\textgreater\footnote{\textless summary\textgreater~is placed before \textless think\textgreater~to enable explicit supervision on the summary segment.} tag in the prompt. 

% \textcolor{red}{Empirical results in Fig.~\ref{fig:tag} report training-free, inference-only structural prompting performance, and are used solely to select an appropriate structural format prior to RL optimization. While introducing explicit structural tags may initially reduce accuracy without training, the purpose of this comparison is to identify a learnable structure for subsequent policy optimization.}

Specifically, for an input $x = (\text{Video}, \text{Question})$, the model $\pi$ is used to generate a group of $G$ sampled outputs:
\begin{equation}
\{\mathcal{O}_g\}_{g=1}^G \sim \pi(\cdot | x) \ .
\end{equation}

% Each individual output $\mathcal{O}_g$ is a sequence of token logits:
% \begin{equation}
%     \mathcal{O}_g = \{o_{g,t}\}_{t=1}^T, \quad o_{g,t} \in \mathbb{R}^V
% \end{equation}
% where $V$ denotes the vocabulary size and $o_{g,t}$ represents the unnormalized logit vector of the $t$-th output token. Since the output follows our proposed three-part trajectory structure (Summarize, Think, Answer), we segment each sequence $\mathcal{O}_g$ into three contiguous parts for clarity in the subsequent discussion:$$\mathcal{O}_g = \{S_g, P_g, Y_g\}$$

Each individual output $\mathcal{O}_g$ is a sequence of token logits, which follows the three-part structured trajectory (Summarize, Think, Answer). We segment the sequence into contiguous parts for clarity in the subsequent discussion:
\begin{equation}
    \mathcal{O}_g = \{o_{g,t}\}_{t=1}^T, \quad \text{where } \mathcal{O}_g = \{S_g, P_g, Y_g\},
\end{equation}
where $V$ denotes the vocabulary size, $o_{g,t} \in \mathbb{R}^V$ represents the unnormalized logit vector of the $t$-th output token, and $\mathcal{O}_g$ is segmented into the Summary ($S_g$), Thinking ($P_g$), and Answer ($Y_g$) segments.
Each segment are defined as:
\begin{equation}
\begin{aligned}
\textbf{Summary: } 
S_g &= \{s_{g,t}\}_{t=1}^{T^s} = \{o_{g,t}\}_{t=1}^{T'}, \\
\textbf{Thinking: } 
P_g &= \{p_{g,t}\}_{t=1}^{T^p} = \{o_{g,t}\}_{t=T'+1}^{T''}, \\
\textbf{Answer: } 
Y_g &= \{y_{g,t}\}_{t=1}^{T^y} = \{o_{g,t}\}_{t=T''+1}^{T}.
\end{aligned}
\end{equation}

Here, $T$ denotes the total number of output tokens. $T'$, $T''$ are the boundary indices that separate the summary, thinking, and answer segments, respectively, corresponding to the sequential token indices $[1, T']$, $[T'+1, T'']$, and $[T''+1, T]$.

% Through this structure, the model is guided to first summarize key actions and events occurring in the video, capturing both semantic and temporal order information, before entering the reasoning stage. 
% This design effectively establishes a top-down reasoning paradigm: the model first achieves global comprehension of the video (top-level abstraction), and then progressively reasons toward the question-specific answer.

\subsection{Consistency of Vision Knowledge (CVK)}
\label{sec:cvk}

\textbf{Group Consistency of Summarization in RL.} 
The generated summaries are not always tightly grounded in the visual content. Instead of relying on an SFT process, we directly introduce a structural supervision signal within the RL framework to enforce visual fidelity in the summary.

% This approach is based on a critical assumption: since the underlying visual content is fixed and factual, its high-level action summary should exhibit strong consistency (low entropy) across different generation attempts. Formally, for a fixed input $x$, all sampled summaries $S_g$ should be drawn from a highly concentrated conditional distribution $\mathcal{D}_S$:

% $$\forall g \in [1, G], \quad S_g \sim \mathcal{D}_S(\cdot | x)$$

% where $\mathcal{D}_S$ represents a distribution tightly aligned with a singular, factual semantic anchor.

% \begin{hypothesis}[Factual Consistency Principle]
% The underlying visual content of a video is \textit{fixed and factual}. Consequently, for a given input $x$, any set of sampled high-level action summaries, $\{S_g\}_{g=1}^G$, generated by a robust model should exhibit \textbf{strong semantic consistency} across different generations. Formally, all sampled summaries $S_g$ must be drawn from a highly concentrated distribution $\mathcal{D}_S$:
% \begin{equation}
%  \forall g \in [1, G], \quad S_g \sim \mathcal{D}_S(\cdot | x)   
% \end{equation}
% where $\mathcal{D}_S$ represents a distribution tightly aligned with a singular, factual semantic anchor.
% \end{hypothesis}

We begin with the assumption that the underlying visual content of a video is \textit{fixed and factual}. Therefore, for a given input $x$, any set of sampled high-level action summaries, $\{S_g\}_{g=1}^G$, generated by a robust model should exhibit \textbf{strong semantic consistency} across different generations. Formally, all sampled summaries $S_g$ must be drawn from a highly concentrated distribution $\mathcal{D}_S$:
\begin{equation}
 \forall g \in [1, G], \quad S_g \sim \mathcal{D}_S(\cdot | x)  \quad, 
\end{equation}
where $\mathcal{D}_S$ represents a distribution tightly aligned with a singular, factual semantic anchor.

Based on this assumption, we propose a group-level consistency objective which enforces summary alignment across the $G$ outputs generated from the same input. Specifically, this objective aims to minimize the semantic dispersion of all sampled summaries $S_g$ around a common consistency anchor $\hat{S}$, which represents the ideal factual summary for the group.

To this end, the objective function must be formulated to either maximize the average similarity or minimize the average consistency cost (dissimilarity) among the group members relative to $\hat{S}$. We integrate this objective within the GRPO framework, as it naturally provides a pipeline to sample $G$ responses for each video–question pair, enabling the direct computation of group-level metrics. It needs two key factors: (1) The Consistency Anchor ($\hat{S}$): This serves as the reference representation that guides the alignment among all group members. (2) The Similarity/Dissimilarity Metric: This measures how closely each generated summary aligns with the anchor in the semantic space.

%%-------

% Based on this assumption, we propose a group-level consistency objective which enforces summary alignment across the $G$ multiple outputs generated from the same input. Specifically, we aim to maximize the semantic similarity of all sampled summaries $S_g$ to a common consistency anchor $\hat{S}$, which represents the ideal factual summary for the group. We formalize this objective as the Summary Consistency Loss (SCL), defined by maximizing the average similarity within the group:
% \begin{equation}
%     \mathcal{L}_{\text{SCL}}(x) = \max_{\pi} \frac{1}{G} \sum_{g=1}^{G} \text{Sim}(S_g, \hat{S})
% \end{equation}

% We implement this objective within the Group Relative Policy Optimization (GRPO) framework, as it naturally provides a pipeline to sample $G$ responses for each video–question pair, enabling the direct computation of group-level metrics. The implementation of $\mathcal{L}_{\text{SCL}}$ requires the definition of two key factors: (1)The Consistency Anchor ($\hat{S}$): This serves as the reference representation that guides the alignment among all group members. (2)The Similarity Metric ($\text{Sim}(\cdot, \cdot)$): This measures how closely each generated summary aligns with the anchor in the semantic space.

\paragraph{GT Supervised Summary Alignment.} 

When ground-truth (GT) summaries $S^{\text{gt}}$ are available, we use them as the definitive consistency anchor, i.e., $\hat{S} = S^{\text{gt}}$. To accurately assess the fidelity of each generated summary, we employ a composite similarity metric that integrates both semantic and lexical signals via sBERT and BLEU scores, respectively. 
% This joint metric enables the model to account for high-level semantic alignment while retaining sensitivity to surface-level textual correspondence. 
Formally, the similarity between the $g$-th sampled summary $S_g$ and the GT anchor is defined as:
\begin{equation}
 \text{Sim}(S_g, \hat{S}) = \alpha \cdot \text{sBERT}(S_g, S^{\text{gt}}) + \beta \cdot \text{BLEU}(S_g, S^{\text{gt}}),
\end{equation}
where $\alpha$ and $\beta$ are weights to balance the two components and ensure the score lies in $[0, 1]$.

We incorporate this similarity score into the RL objective by augmenting the original answer-based reward $R_g$ with a summary-consistency reward, resulting in:
\begin{equation}
R_g' = \gamma_1 R_g + \gamma_2 \text{Sim}(S_g, \hat{S})\ ,
\end{equation}
where $\gamma_1$ and $\gamma_2$ are scaling factors.

\vspace{1mm}\noindent\textbf{Self-Supervised Summary Alignment.} Obtaining GT summaries is costly and time-consuming, and strict alignment to GT annotations may constrain the model’s expressive capacity or induce overfitting and bias. Therefore, we extend the group consistency objective to a self-supervised setting that eliminates the need for manual annotations. We dynamically derive the consistency anchor ($\hat{S}$) from the model's own predictions and identify high-quality summaries based on the binary correctness indicator $R_g^C$:
\begin{equation}
R_g^C =
\begin{cases}
1, & Z_g = Z^{\text{gt}}, \\
0, & Z_g \neq Z^{\text{gt}},
\end{cases}
\end{equation}
where $Z_g$ and the $Z^{gt}$ denote the predicted and ground-truth answers, respectively. The position-wise center $S^C$, which serves as our consistency anchor $\hat{S}$, is then computed by aggregating the token representations of all selected high-quality summaries. Let $S^C = \{s_t^C\}_{t=1}^{T^s}$ denote the consistency center, and $S_g = \{s_{g,t}\}_{t=1}^{T^s}$ denote the $g^{th}$ summary's token representation. For each position $t = 1, 2, \ldots, T^s$, the position-wise center $s_t^C$ is computed as:
\begin{equation}
\label{eq:anchor}
    s_t^C = \frac{1}{\sum_{g=1}^{G} R_g^C} \sum_{g=1}^G R_g^C \cdot s_{g,t}\ .
\end{equation}
% $$s_t^C = \frac{1}{\sum_{g=1}^{G} R_g^C} \sum_{g=1}^G R_g^C \cdot s_{g,t}\ .$$

To implement the self-supervised objective, we utilize Kullback–Leibler (KL) divergence as our dissimilarity metric to quantify the position-wise inconsistency within the group. The KL divergence $\mathcal{D}_t$ at position $t$ is calculated as:
\begin{equation}
    \mathcal{D}_t = \frac{1}{G} \sum_{g=1}^{G} D_{\text{KL}}(s_{g,t} || s^C_{t})\ .
\end{equation}
% $$\mathcal{D}_t = \frac{1}{G} \sum_{g=1}^{G} D_{\text{KL}}(s_{g,t} || s^C_{t})\ .$$
To align with the GRPO Policy Objective, we convert this inconsistency measure into the Summary Token Weight ($\omega_t^S$). Since a larger divergence ($\mathcal{D}_t$) indicates lower consistency (less agreement), the policy should assign a smaller weight to that token position. This encourages the model to focus on learning stable and consistent parts of the summary. We define the Summary Token Weight $\omega_t^S$ as: 
\begin{equation}
    \omega_t^S = 1 - \lambda \cdot \frac{\mathcal{D}_t - \mathcal{D}_{\min}}{\mathcal{D}_{\max} - \mathcal{D}_{\min}}\ .
\end{equation}
% $$\omega_t^S = 1 - \lambda \cdot \frac{\mathcal{D}_t - \mathcal{D}_{\min}}{\mathcal{D}_{\max} - \mathcal{D}_{\min}}\ .$$
%
Here, $\lambda \in [0, 1]$ controls the scaling intensity. This weight $\omega_t^S$ is then applied to the Summary segment ($1 \le t \le T'$) of the total GRPO Policy Objective, as illustrated in $\mathcal{J}_{\text{total}}(\theta)$, to realize the Self-Supervised Consistency goal.

\subsection{Dynamic Variety of Reasoning (DVR) }
\label{sec:dvr}
While summary consistency stabilizes the factual grounding of the model, excessive uniformity in subsequent reasoning paths can be detrimental.  Therefore, we introduce the Dynamic Variety of Reasoning (DVR) objective to encourage diversity in the \textless think\textgreater~stage of structured CoT.

Specifically, we encourage diversity in the subsequent Thinking segment ($P$) by focusing on the entropy of the token distribution at each position. The Diversity Weight $\omega^d_{g,t}$ is directly proportional to the measured entropy $H_{g,t}$, which is calculated over the predicted token distribution $p_{g,t}$:
\begin{equation}
    H_{g,t} = -\, p_{g,t}^{\top} \log p_{g,t}.
\end{equation}
% $$H_{g,t} = -\, p_{g,t}^{\top} \log p_{g,t}.$$ 
This measured entropy is then normalized to define the base Diversity Weight $\omega^d_{g,t}$:
\begin{equation}
    \begin{aligned}
\omega^d_{g,t} &= 1 + \lambda' \cdot \frac{H_{g,t}-H_{\min}}{H_{\max}-H_{\min}}, \\
H_{\min} &= \min_{g, t}\{H_{g,t}\}, \quad H_{\max} = \max_{g, t}\{H_{g,t}\} \ ,
\end{aligned}
\end{equation}
% $$\begin{aligned}
% \omega^d_{g,t} &= 1 + \lambda' \cdot \frac{H_{g,t}-H_{\min}}{H_{\max}-H_{\min}}, \\
% H_{\min} &= \min_{g, t}\{H_{g,t}\}, \quad H_{\max} = \max_{g, t}\{H_{g,t}\} \ ,
% \end{aligned}$$
where $\lambda'$ is a scaling hyperparameter.

Simply maximizing diversity can be detrimental to performance as the policy converges. When a group yields a high number of positive samples, it indicates the existing reasoning paths are effective and less exploration is needed. In such high-accuracy groups, excessive diversity may introduce noise.Therefore, we introduce a dynamic modulation coefficient based on the group's performance to adjust the diversity incentive. We define the group accuracy $\mathcal{A}$ as the fraction of correct answers:
\begin{equation}
    \mathcal{A} = \frac{\sum_{g=1}^{G} \mathbf{1}_{Z_g = Z^{\text{gt}}}}{G}.
\end{equation}
% $$\mathcal{A} = \frac{\sum_{g=1}^{G} \mathbf{1}_{Z_g = Z^{\text{gt}}}}{G}.$$
The base Diversity Weight $\omega^d_{i,t}$ is then reweighted by the factor $(1 - \mathcal{A})$ to yield the Dynamic Diversity Weight $\omega^{d'}_{i,t}$:
\begin{equation}
    \omega^{d'}_{i,t} = \omega^{d}_{i,t} \cdot (1 - \mathcal{A}).
\end{equation}
% $$\omega^{d'}_{i,t} = \omega^{d}_{i,t} \cdot (1 - \mathcal{A}).$$
This formulation ensures that the diversity incentive is strongest for groups with low overall accuracy. Conversely, it is minimized for highly accurate groups, preserving stable reasoning paths. This dynamic weight $\omega^{d'}_{i,t}$ is used to inject the diversity signal into the final policy objective $\mathcal{J}_{\text{total}}(\theta)$.

\subsection{Structured Policy Objective}
\label{sec:policy}

We define the final policy objective $\mathcal{J}_{\text{total}}(\theta)$ as maximizing the expected augmented reward, which is integrated with our structural consistency and diversity constraints and the GRPO-specific policy regularization terms.The overall objective is to maximize the following expression:
\begin{equation}
    \mathcal{J}_{\text{total}}(\theta) = \mathcal{J}_{\text{grpo}}^{\text{SCoT}}(\theta) - \mathcal{J}_{\text{reg}}(\theta) \ .
\end{equation}
% $$\mathcal{J}_{\text{total}}(\theta) = \mathcal{J}_{\text{grpo}}^{\text{SCoT}}(\theta) - \mathcal{J}_{\text{reg}}(\theta) \ .$$

The structured GRPO objective $\mathcal{J}_{\text{grpo}}^{\text{SCoT}}(\theta)$ is designed to maximize the token-wise weighted advantage, the weights $W_{g,t}$ enforce the consistency constraints in the summary and dynamic diverstiy in the thinking.
This objective is defined as:
\begin{equation}
\begin{aligned}
\mathcal{J}_{\text{grpo}}^{\text{SCoT}}(\theta) = & \mathbb{E} \Big[ \frac{1}{G} \sum_{g=1}^{G} \big( \sum_{t=1}^{T} W_{g,t} \cdot \min \left( r_{g,t}(\theta) \cdot A_{g,t}, \right.   \\
& \left. \quad  \text{clip}(r_{g,t}(\theta), 1-\epsilon, 1+\epsilon) \cdot A_{g,t} \right) \big) \Big] \ , \\
r_{g,t}(\theta) &= \frac{\pi_{\theta}(o_{g,t}|q, o_{g,<t})}{\pi_{\text{old}}(o_{g,t}|q, o_{g,<t})}.
\end{aligned}
\end{equation}

%$$r_{g,t}(\theta) = \frac{\pi_{\theta}(o_{g,t}|q, o_{g,<t})}{\pi_{\text{old}}(o_{g,t}|q, o_{g,<t})}$$

The term $A_{g,t}$ represents the overall relative advantage (computed by mean-variance normalization of the enhanced reward $R'_g$), and the Token-Wise Weights $W_{g,t}$ modulate the policy update strength at each token position:
\begin{equation}
    W_{g,t} =
\begin{cases}
\omega_t^S\ , & 1 \le t \le T'  \\
\omega_{g,t}^{d'}\ , & T'+1 \le t \le T 
\end{cases}
\end{equation}
% $$W_{g,t} =
% \begin{cases}
% \omega_t^S\ , & 1 \le t \le T'  \\
% \omega_{g,t}^{d'}\ , & T'+1 \le t \le T 
% \end{cases}$$

\subsection{Dataset Construction}
% \begin{figure}[!t]
%     \centering
%     \includegraphics[width=1\linewidth]{new_data2.pdf}
%     \caption{Data distribution of the EventFlowQA-80K dataset.
% The dataset consists of two hierarchical levels: Action-level questions and Video-level questions. The outer ring shows the detailed proportions of each temporal reasoning category, including action order, duration, comparison, causality, anticipation, and others.}
%     \label{fig:data_dis}
% \end{figure}

Previous work has consistently highlighted a critical limitation in MLLMs: the inability to robustly capture and utilize fine-grained temporal information~\cite{feng2025breaking}. To facilitate effective video understanding training, which is a necessity for our proposed SDRL framework, we introduce EventFlowQA, a comprehensive video question-answering dataset focused on intricate action sequencing and temporal causality.
In total, EventFlowQA comprises 53K high-quality QA pairs (50K for training and 3K for validation). The final distribution is over 15 focused temporal aspects,
%is illustrated in Figure~\ref{fig:data_dis}, 
serving as the core benchmark for all ablation studies. The detailed methodology for dataset construction and analysis is provided in the Supplementary Material. 
\section{Experiment}
\begin{table}[t]
    \centering
        \caption{Ablation study of CVK and DVR modules on the EventFlowQA.
Rows (a–g) use GT supervision, while (h–l) adopt self-supervision.
Adding semantic (sBERT, BLEU) and distributional (KL, Entropy, Dynamic) constraints improves accuracy.}%, showing the effectiveness of each component and the benefit of self-supervised consistency.}
    \resizebox{0.98\linewidth}{!}{
    \begin{tabular}{c|cc|ccc|c}
    \toprule
         & \multicolumn{2}{|c|}{\textbf{CVK}} &    \multicolumn{3}{|c|}{\textbf{DVR}} & \multirow{2}{*}{\textbf{Accuracy}} \\ \cline{2-6}
          & \textbf{sBERT} & \textbf{BLEU} & \textbf{KL} & \textbf{Entropy} & \textbf{Dynamic} & \\ \midrule
         Orig. & - & - & - & - & -& 42.37  \\
        \midrule
        (a) & & \checkmark & & & &43.85 \\
         (b) & \checkmark & & & & & 46.32\\
         (c) & \checkmark & \checkmark& & & & 48.56 \\
         (d) & \checkmark & \checkmark& \checkmark& & &  46.71\\
         (e) & \checkmark & \checkmark& \checkmark& & \checkmark&  49.13\\
         (f) & \checkmark & \checkmark& & \checkmark& &  50.09\\
         (g) & \checkmark & \checkmark& & \checkmark& \checkmark& 52.22\\
         \midrule
         (h) & \multicolumn{2}{c|}{self supervision} & & & & 54.28 \\
         (i) & \multicolumn{2}{c|}{self supervision} & \checkmark& &  & 53.34 \\
         (j) & \multicolumn{2}{c|}{self supervision} &\checkmark & &\checkmark & 55.78\\
         (k) & \multicolumn{2}{c|}{self supervision} & & \checkmark &  &  54.13\\
         (l) & \multicolumn{2}{c|}{self supervision} & & \checkmark & \checkmark & 56.10 \\
         \bottomrule
    \end{tabular}}
    \label{tab:ablation}
\end{table}

\begin{table*}[t]
    \centering
        \caption{Evaluation of Video MLLMs showing performance (Accuracy \%) across various benchmarks, categorized by their training strategy (SFT, RL, SFT+RL). Our model achieves state-of-the-art results by employing the RL-only strategy, demonstrating superior performance across most metrics. `Ours\textsuperscript{†}' denotes the variant trained using the EventFlow dataset. VideoRFT\textsuperscript{*} indicates evaluation with 16-frame input.}
      \resizebox{\linewidth}{!}{
    \begin{tabular}{c|c|llll|lll}
    \toprule
         \multirow{2}{*}{\textbf{Models}}&  \multirow{2}{*}{\textbf{Training}} &  \multicolumn{4}{|c|}{\textbf{Video Reasoning Benchmark} }& \multicolumn{3}{|c}{\textbf{Video General Benchmark}}\\ \cline{3-9}
         & & \textbf{NExT-GQA} & \textbf{MMVU} & \textbf{VideoMMMU}& \textbf{VSIBench}& \textbf{MVBench} & \textbf{TempCompass} & \textbf{VideoMME} \\ \midrule
         LLaMA-VID~\cite{li2024llamavid} & \multirow{11}{*}{None}& -& - & - & - & 41.9 & 45.6 & -\\ 
         VideoLLaMA2~\cite{cheng2024videollama2}& & - & 44.8 & - & -& 54.6 & - & 47.9 \\
         LongVA-7B~\cite{zhang2024longva} &  & -& - & 23.9 & 29.2& - & 56.9 & 52.6\\
         VILA-1.5-8B~\cite{lin2024vila} & & -& - & 20.8 & 28.9 & - & 58.8 & - \\
         % VILA-1.5-40B ~\cite{lin2024vila}& & -& - & 34.0  & 31.2& - & - & 60.1\\
         Video-UTR-7B~\cite{yu2025videoutr} &  & -& - & -& - &58.8 & 59.7 & 52.6\\
         LLaVA-OneVision-7B~\cite{li2024llava} &  & -& 49.2 & 33.8 &32.4 & 56.7 & - & \textbf{58.2}\\
         Kangaroo-8B~\cite{liu2024kangaroo} &  & -& - & -&- &61.1& 62.5 & 56.0\\
         Qwen2.5-VL-7B ~\cite{bai2025qwen25vl} & & 75.9 & 65.4 &48.4 & 29.1& 63.3 & 72.5 & 56.5\\
         Qwen2.5-VL-7B (video-r1 CoT)~\cite{bai2025qwen25vl} & & -& 59.2 & 47.8 & 27.7 & 57.4 & 72.2 & 53.1 \\
         Qwen2.5-VL-7B (ours CoT)~\cite{bai2025qwen25vl} &   & 73.6 & 63.2 & 49.3 & 26.6 & 58.9 & 69.5 & 49.0\\ \midrule
         % Qwen2.5-VL-7B~\cite{bai2025qwen25vl} & SFT  & \\
         Video-R1~\cite{feng2025videor1} & SFT & - & 51.3 & 47.4 & 31.8  & 59.4 & 69.2 & 52.8\\
         VideoRFT\textsuperscript{*}~\cite{wang2025videorft} & SFT & -& 60.5 & 48.5 & 31.7 & 57.0 & 68.4 & 54.1 \\
         \midrule
         Video-R1~\cite{feng2025videor1} &  SFT+ RL  &  74.3 &64.2 & \textbf{52.4} & 34.6 & 62.7 & 72.6 & 57.4\\ 
         VideoRFT\textsuperscript{*}~\cite{wang2025videorft} & SFT+ RL & 75.1 & 67.3 & 50.6& 35.7 & 61.4 & 73.1 & 58.1 \\ \midrule
         Video-R1~\cite{feng2025videor1} & RL & -& 63.8 & 49.5 & 31.8 & 60.4 & 70.9 & 53.8\\
         VideoChat-R1~\cite{li2025videochatr1} &   RL & 76.0 & 64.2 & -& -& 63.1  & 72.9 & 52.4\\
         VideoRFT\textsuperscript{*}~\cite{wang2025videorft} & RL & -& 63.5 & 47.4 & 32.1 & 59.2 & 70.8 & 51.9 \\ 
         TW-GRPO~\cite{dang2025twgrpo}&  RL &  76.1 & 65.8 & -&- & 63.3 & 73.3 & 55.1\\ 
         \rowcolor[HTML]{E6F4EA}
         SDRL (Ours\textsuperscript{†}) &  RL  & 77.3 {\textcolor[HTML]{006400}{ (+3.7)}}& 64.8 {\textcolor[HTML]{006400}{ (+1.6)}} & 51.1 {\textcolor[HTML]{006400}{ (+1.8)}} & \textbf{36.1} {\textcolor[HTML]{006400}{ (+9.5)}} & 63.3 {\textcolor[HTML]{006400}{ (+4.4)}} & \textbf{74.4} {\textcolor[HTML]{006400}{ (+4.9)}} & 53.1 {\textcolor[HTML]{006400}{ (+4.1)}}  \\
         \rowcolor[HTML]{E6F4EA}
         SDRL (Ours) & RL & \textbf{79.3} {\textcolor[HTML]{006400}{ (+5.7)}} & \textbf{68.6} {\textcolor[HTML]{006400}{ (+5.4)}} & 51.3 {\textcolor[HTML]{006400}{ (+2.0)}}&  32.9 {\textcolor[HTML]{006400}{ (+6.3)}} & \textbf{64.2} {\textcolor[HTML]{006400}{ (+5.3)}} & 73.4 {\textcolor[HTML]{006400}{(+3.9)}}& 54.7 {\textcolor[HTML]{006400}{ (+5.7)}}
\\\bottomrule
    \end{tabular}}
    \label{tab:sota}
\end{table*}
% \footnote{We evaluate VideoRFT with 16 frames for fair comparison of all the methods.}

\textbf{Training Setup.} 
% We adopt an RL-only training paradigm to isolate the effect of policy optimization from SFT. The model is trained on the Video-R1-260K dataset (video-only subset)and the proposed EventFlowQA dataset.
We adopt an RL-only training paradigm to isolate the effect of policy optimization from SFT.
We use Qwen2.5-VL-Instruct-7B as the backbone. Each training sample consists of 16 uniformly sampled frames at a resolution of 128×28×28, and inference is conducted under the same 16-frame setting for consistency.
For the ground-truth–supervised consistency objective, we set weighting coefficients to $\alpha=0.7$ and $\beta=0.3$, with $\gamma_1=1$ and $\gamma_2=1$. For the self-supervised CVK and DVR objectives, we use $\lambda=0.5$ and $\lambda'=0.7$. Training is performed on 32 NVIDIA A100 GPUs with a GRPO group size of 8 for a total of 1,000 RL iterations.
% For the ablation study, we train models on EventFlowQA, which includes ground-truth summaries with temporally ordered action annotations. All other configurations remain consistent with the Video-R1-260K setup, enabling a complete comparison between ground-truth supervision and self-supervised consistency learning. 

We conduct two types of experiments in this paper. 
(1) All ablation studies are conducted on our proposed EventFlowQA dataset, which provides ground-truth action sequence annotations and enables controlled analysis of both GT-supervised and self-supervised summary consistency and diversity mechanisms.
(2) Benchmark comparisons (Table~\ref{tab:sota}) are conducted on seven public VideoQA benchmarks. For fair comparison, SDRL is trained on the same Video-R1-260K training data as prior RL-based methods, unless otherwise specified (e.g., Ours$^\dagger$ trained on EventFlowQA).
 Further implementation details are provided in the supplementary material.

\vspace{1mm}\noindent\textbf{Benchmarks.} 
We evaluate our model comprehensively across seven widely used video understanding benchmarks: NExT-GQA~\cite{xiao2024can}, MMVU~\cite{zhao2025mmvu}, VideoMMMU~\cite{hu2501video}, VSIBench~\cite{yang2025thinking}, MVBench~\cite{li2024mvbench}, TempCompass~\cite{liu2024tempcompass}, and VideoMME~\cite{fu2025video}.
These benchmarks can be broadly categorized into two groups: (1) \textit{Video Reasoning Benchmarks} (NExT-GQA, MMVU, VideoMMMU, VSIBench) are designed to assess a model’s temporal and causal reasoning capabilities, including multi-choice question answering, compositional inference, and long-range dependency understanding. (2) \textit{General Video Understanding Benchmarks} (MVBench, TempCompass, VideoMME) focus on holistic video comprehension, integrating perception-level understanding (e.g., object, action, and event recognition) with high-level reasoning abilities.

% We comprehensively evaluate our model on seven widely used video benchmarks: NExT-GQA~\cite{xiao2024can}, MMVU~\cite{zhao2025mmvu}, VideoMMMU~\cite{hu2501video}, VSIBench~\cite{yang2025thinking}, MVBench~\cite{li2024mvbench}, TempCompass~\cite{liu2024tempcompass}, and VideoMME~\cite{fu2025video}.
% Video Reasoning Benchmarks (NExT-GQA, MMVU, VideoMMMU, VSIBench), designed to test a model’s capability for temporal and causal reasoning, including multi-hop question answering, compositional inference, and long-range dependency understanding.
% General Video Understanding Benchmarks(MVBench, TempCompass, VideoMME), focusing on holistic video comprehension, which integrates perception-level understanding (e.g., object, action, and event recognition) with high-level reasoning ability.

% This comprehensive benchmark suite allows us to evaluate not only reasoning capability but also generalization and robustness across diverse video understanding tasks.

\subsection{Comparision with State of the Art method}

Table~\ref{tab:sota} presents a comprehensive comparison between our method and recent Video MLLMs across both video reasoning and general understanding benchmarks. Overall, our RL-only framework (Ours) achieves consistent state-of-the-art performance, surpassing both SFT-only and SFT+RL pipelines.
% On reasoning benchmarks such as NExT-GQA, MMVU, and VideoMMMU, our model outperforms the SFT+RL method (VideoRFT*) by +4.2\%, +1.3\%, and +0.7\%, respectively. On general benchmarks including MVBench and TempCompass, SDRL further yields gains of +2.8\% , +0.3\%, demonstrating strong generalization.
On reasoning benchmarks such as NExT-GQA, MMVU, and VideoMMMU, our model surpasses the SFT+RL baseline (VideoRFT*) by 4.2\%, 1.3\%, and 0.7\%, respectively. Furthermore, SDRL delivers consistent gains of 2.8\% and 0.3\% on general benchmarks like MVBench and TempCompass, validating its robust generalization across diverse tasks.
Moreover, compared to other single-stage RL methods (e.g., VideoChat-R1, TW-GRPO), SDRL consistently achieves higher accuracy across all metrics. When compared with the base model Qwen2.5-VL, our approach yields accuracy improvements of up to +6.3\% and +9.5\% points on VSIBench (as indicated by the green numbers), even under different training data settings. 
% Using EventFlowQA, which is just 20\% the size of the Video-R1 training dataset, SDRL achieves the highest performance on TempCompass, highlighting the efficiency of our proposed dataset. 
By training on EventFlowQA, which is only 20\% the size of the Video-R1 RL training set, SDRL achieves superior performance on TempCompass, underscoring the high data efficiency of our proposed dataset.
Distinct from previous approaches that rely on SFT followed by RL fine-tuning, our RL-only strategy stabilizes optimization through structured reasoning supervision, delivering both higher accuracy and greater training efficiency.

\subsection{Analysis in Consistency of Summarization}

\textbf{Ground-Truth Supervision vs.\ Self-Supervision.}
% Although ground-truth (GT) supervision intuitively provides stronger guidance by anchoring predictions to human references, our experiments reveal a more nuanced trend driven by model scale and pre-training. As shown in Table~\ref{tab:ablation}, with the larger 7B model, the self-supervised approach achieves a greater accuracy gain than GT supervision ($+11.91$ vs. $+6.19$). We attribute this to the catastrophic forgetting risk in large, highly pre-trained models. Naive, strict GT supervision, anchored to a small set of human summaries, over-constrains the optimization, potentially erasing valuable global semantic knowledge acquired during pre-training. In contrast, self-supervision (via CVK) leverages semantic consistency among model-generated samples, refining temporal and factual reasoning without destructive parameter shifts.
% This effect reverses with the smaller 3B model in Table~\ref{tab:consis_size}, where GT supervision provides a slightly better gain. This confirms that smaller models rely more on external GT guidance, while larger models benefit from self-consistency–based regularization. 
Although GT supervision intuitively provides stronger guidance by anchoring predictions to human references, our results reveal a nuanced trend influenced by model scale and pre-training. As shown in Table~\ref{tab:ablation}, the larger 7B model benefits more from self-supervision ($+11.91\%$) than from GT supervision ($+6.19\%$), likely due to catastrophic forgetting that strict GT alignment with limited human summaries can over-constrain optimization and suppress useful semantic priors from pre-training. In contrast, self-supervision exploits semantic consistency among predictions, enhancing temporal and factual reasoning without destabilizing parameters.
For the smaller 3B model (Table~\ref{tab:consis_size}), GT supervision yields slightly higher gains, indicating that smaller models depend more on explicit human guidance, whereas larger ones benefit from self-consistency regularization.
\begin{table}[t]
    \centering
        \caption{Effect of model size under different supervision types (Accuracy \%). Smaller model gains more under GT supervision.}
    \resizebox{0.8\linewidth}{!}{
    \begin{tabular}{c|c|cc}
    \toprule
        \textbf{ Model size}& \textbf{Original} & \textbf{GT sup.}& \textbf{Self sup.}\\ \midrule
        3B & 41.46 & 44.47\,{\textcolor[HTML]{006400}{ (+3.01)}} & 43.86 {\textcolor[HTML]{006400}{(+2.40)~}}  \\
        7B & 42.37 & 48.56 {\textcolor[HTML]{006400}{(+6.19)}}  & 54.28 {\textcolor[HTML]{006400}{(+11.91)}}  \\
         \bottomrule
    \end{tabular}}
    \label{tab:consis_size}
    \vspace{-1mm}
\end{table}

\begin{table}[t]
\centering
\caption{Comparison of BLEU and sBERT scores based on CVK.}
  \resizebox{0.8\linewidth}{!}{
\begin{tabular}{l|ccc}
\toprule
 & \textbf{Accuracy} (\%)& \textbf{BLEU (\%)} & \textbf{ sBERT (\%)} \\
\midrule
w/o & 42.37 & 8.84 & 70.33 \\
w & 54.28 & 12.57  & 79.76 \\
\bottomrule
\end{tabular}}
\label{tab:cvk}
\end{table}

% -----v0
% \paragraph{Which metric is most effective for enforcing the consistency constraint under GT supervision?} 
% To identify the most effective metric for enforcing the consistency constraint under GT supervision, we compare BLEU and sBERT, which measure similarity between predicted and ground-truth summaries from different perspectives. BLEU focuses on lexical overlap through n-gram precision, while sBERT captures semantic alignment in the embedding space. As shown in Table 2, using sBERT alone (a) achieves a notable improvement over BLEU alone (0), 46.32\% vs. 43.85\%, demonstrating that semantic-level supervision provides stronger and more stable guidance than surface-level token matching.

% Interestingly, combining the two metrics yields further gains: the joint BLEU + sBERT setting (c) reaches 46.71\%, exceeding either metric individually. This suggests that BLEU and sBERT contribute complementary information—BLEU helps maintain grammatical fidelity and local fluency, while sBERT ensures global semantic coherence. The combination thus provides a more balanced and fine-grained supervision signal, aligning generated summaries with both linguistic form and meaning.

\begin{figure*}[t]
    \centering
    \includegraphics[width=\linewidth,height=.55\linewidth]{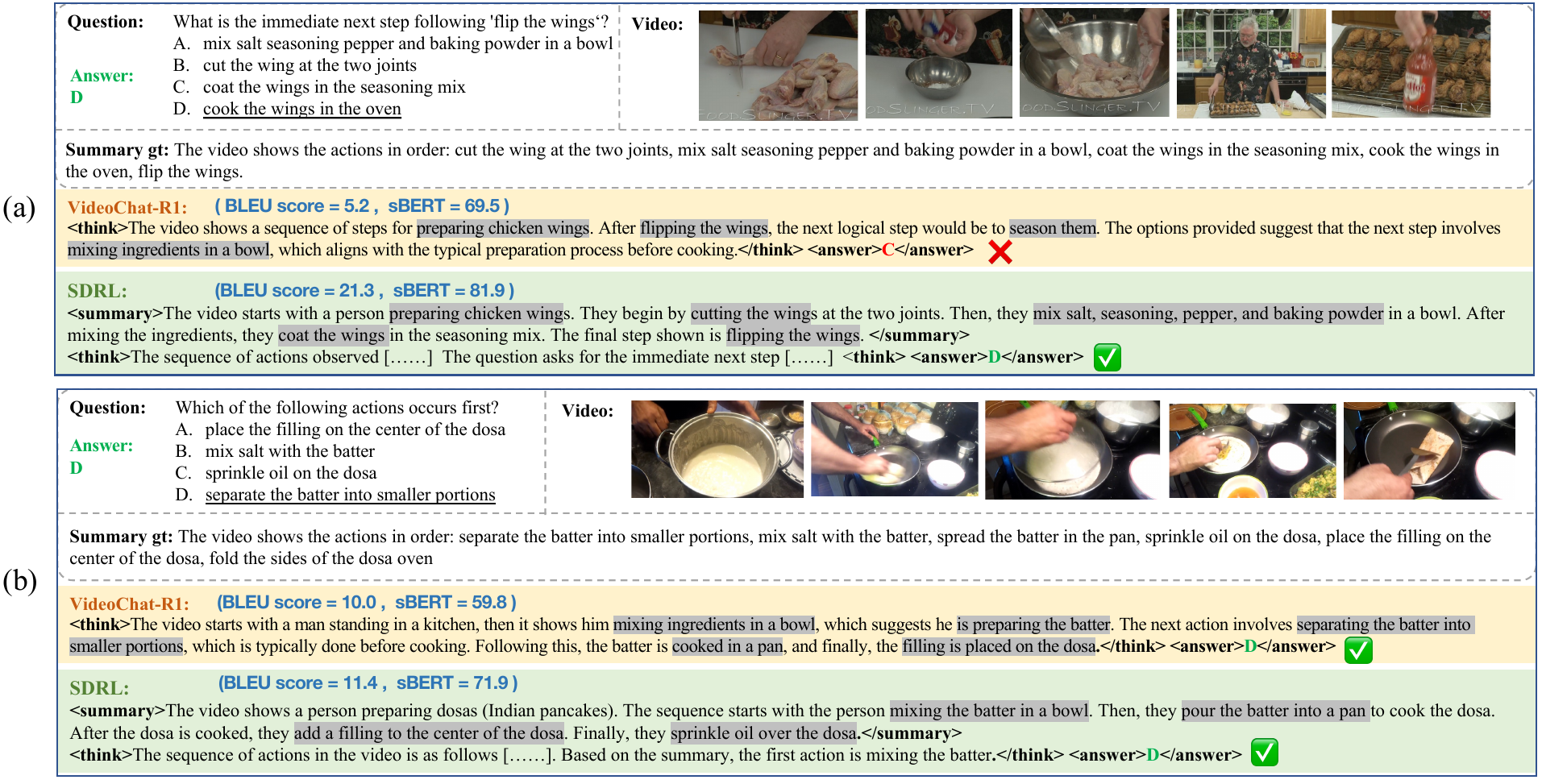}
    \caption{Comparison of CoTs and final answers generated by VideoChat-R1 and our proposed SDRL method. SDRL demonstrates superior grounding and logic flow, evidenced by higher BLEU and sBERT scores relative to the ground truth summary.}
    \label{fig:vis}
    \vspace{-1mm}
\end{figure*}

\vspace{1mm}
\noindent\textbf{Metric for Consistency Constraint under GT Supervision.}
To determine the most effective metric for enforcing summary consistency under ground-truth (GT) supervision, we compare BLEU and sBERT, which evaluate similarity between predicted and GT summaries from complementary perspectives. 
%BLEU captures lexical overlap via n-gram precision, while sBERT measures semantic alignment within the embedding space.
As shown in Table~\ref{tab:ablation}, using (b)sBERT alone outperforms (a) BLEU alone(46.32\% vs. 43.85\%),  indicating that semantic-level supervision provides stronger and more stable guidance than surface-level token matching. Moreover, combining BLEU and sBERT (c) yields the highest accuracy of 46.71\%, surpassing either metric individually. This result suggests that BLEU and sBERT offer complementary benefits, and their integration thus delivers a more balanced supervision signal.
%BLEU preserves grammatical fidelity and local fluency, while sBERT enhances global semantic coherence. 

\subsection{Analysis in the diversity of the thinking}

\textbf{Is diversity necessary within the group?}
As shown in Table~\ref{tab:ablation}, when diversity enhancement is applied statically to all groups, such as in  (d) and (f), the model achieves accuracies of 49.13\% and 55.78\%. 
By dynamically applying diversity to encourag exploration only in uncertain or incorrect groups while maintaining stability in confident ones, the accuracy consistently rises to 52.22\% and 56.10\%.
These results reveal that indiscriminately enforcing diversity can degrade performance when the model is already confident. In contrast, the dynamic strategy \textit{focuses exploration on uncertain groups}, fostering richer reasoning where necessary while \textit{preserving stability} for confident ones. By suppressing diversity in “solved” groups and amplifying it in “unsolved” ones, the model receives stronger, more informative learning signals, leading to higher overall accuracy. 
% This finding underscores that diversity should be treated as a conditional exploration signal rather than a uniform regularization term.

% \vspace{1mm}
\noindent\textbf{Which metric is more effective in promoting diversity?}
% We further explore which metric better promotes reasoning diversity during GRPO training. As shown in Table 2, we compare two diversity-related objectives—KL divergence and entropy regularization—under both GT and self-supervised consistency settings. The results show that entropy consistently yields higher accuracy and more stable performance than KL divergence. For instance, under GT supervision, the model using KL in (d) achieves 49.13\%, while replacing it with entropy in (e) improves accuracy to 50.09\%. A similar trend is observed in the self-supervised setting: (i) with KL reaches 55.78\%, whereas (k) with entropy and dynamic gating further increases accuracy to 56.10\%.
% This performance gap reflects a fundamental difference between the two metrics. KL divergence measures the pairwise discrepancy between tokens across samples, providing \textit{a local and position-dependent constraint}. While this can align token-level distributions, it may inadvertently suppress the global variability of the reasoning process. In contrast, entropy regularization operates as a \textit{global uncertainty measure}, encouraging the model to maintain a balanced exploration of reasoning paths without collapsing to deterministic patterns. This global control allows the model to preserve semantic diversity while still converging toward confident answers.
We further examine which metric better promotes reasoning diversity during GRPO training. As shown in Table~\ref{tab:ablation}, we compare two diversity objectives, \textit{i.e.}, KL divergence and entropy regularization, under both GT and self-supervised consistency settings. Entropy consistently yields higher accuracy and more stable performance than KL. For example, under GT supervision, replacing KL (49.13\%, (e)) with entropy (50.09\%, (f)) improves accuracy, and a similar gain is observed in the self-supervised setting (55.78\% → 56.10\%).
This performance gap arises from their intrinsic difference: KL divergence enforces \textit{local, position-dependent alignment} across token distributions, which can suppress global variability, while entropy regularization acts as a \textit{global uncertainty control}, encouraging balanced exploration without collapsing into deterministic reasoning. Consequently, entropy better preserves semantic diversity while stabilizing the reasoning process.

% \subsection{Other Analysis}
% \paragraph{Different initialization.}

% \paragraph{Compare GRPO with SFT.} It can be observed that across various types of tasks, GRPO outperforms SFT.
% Whether it is in terms of the performance on in-domain tasks, out-domain tasks, or the preservation
% of the original general performance, our experimental results demonstrate that GRPO is a promising
% fine-tuning approach. We will leave the large-scale comparison for future research.

\section{Visualization}

Figure~\ref{fig:vis} illustrates qualitative comparisons of the generated summaries and reasoning outputs without GT supervision. 
In Figure~\ref{fig:vis}(a), VideoChat-R1 produces an incorrect answer due to poor temporal reasoning. Its generated summary fails to preserve the correct action order, resulting in a lower BLEU score (5.2) and weak alignment with the ground-truth sequence. In contrast, SDRL generates a summary that closely follows the true action sequence and correctly predicts the answer. This indicates that the proposed summary consistency constraint helps the model capture accurate temporal dependencies and maintain coherent action ordering.
Even when both models produce the correct answer in Figure~\ref{fig:vis}(b), SDRL achieves a higher BLEU score, indicating that its generated summary more faithfully reflects the action sequence and provides clearer temporal organization. 
Table~\ref{tab:cvk} further shows that CVK consistently improves both summary consistency and task accuracy.
%validating the effectiveness of group consistency in enhancing factual grounding and temporal reasoning.
\section{Conclusion}
In conclusion, we introduced Summary-Driven Reinforcement Learning (SDRL), a framework that relies policy optimization paradigm for MLLMs in video understanding. By integrating a Structured Chain-of-Thought (Summarize $\rightarrow$ Think $\rightarrow$ Answer) and leveraging self-supervised Token-Wise Weighting (via CVK and DVR), SDRL alleviates the critical challenges of thinking drift and the multi-Stage training pipeline inherent in existing SFT+RL methods. Our framework forces the model's intermediate reasoning  by anchoring it to the explicit Summary segment, enabling robust and interpretable decision-making without reliance on human-annotated CoT data. 
% Extensive experiments across diverse video benchmarks demonstrate that SDRL achieves state-of-the-art performance, validating the efficacy of self-supervised structural constraints within an efficient RL objective. 
We anticipate that this work will pave the way for future research into single-stage, data-efficient, and structurally robust reasoning frameworks for general MLLMs.

% We present Summary-Driven Reinforcement Learning (SDRL), a framework that redefines policy optimization for MLLMs in complex temporal reasoning. By incorporating a Structured CoT and self-supervised Token-Wise Weighting (CVK and DVR), SDRL mitigates thinking drift and removes the costly multi-stage SFT+RL. Anchoring reasoning to the explicit summary segment enables robust and interpretable decision-making without human-annotated CoT data. Experiments across diverse video benchmarks show that SDRL achieves state-of-the-art performance, highlighting the effectiveness of self-supervised structural constraints for efficient and unified RL-based reasoning.

{
    \small
    \bibliographystyle{ieeenat_fullname}
    \bibliography{main}
}

\clearpage
\maketitlesupplementary

\setcounter{section}{0}
\renewcommand{\thesection}{\Alph{section}}

\section{EventFlow Dataset Construction}

\subsection{QAs Generation Pipeline }
The EventFlow dataset is constructed via a three-stage automated pipeline designed to create high-quality, linguistically diverse Question-Answer (QA) pairs for training and evaluating temporal reasoning capabilities in Reinforcement Learning (RL) agents. The pipeline utilizes action annotations from existing large-scale video datasets, employs a systematic set of temporal logic templates, and leverages a Large Language Model (LLM) for linguistic diversification and question instantiation. The three core stages are:
\begin{itemize}
    \item \textbf{Source Data Selection}: Extraction of video segments and their precise $(action, start\_time, end\_time)$triplets.
    \item \textbf{Template Design}: Definition of 15 fundamental temporal logic templates (e.g., precedence, duration, concurrency).
    \item \textbf{LLM Generation}: Using an LLM to rewrite templates for linguistic diversity and subsequently instantiate the questions using real action labels and derive the ground-truth answers based on temporal logic.
\end{itemize}

% \subsubsection{Source Datasets with Temporal Annotations}
% We selected three widely-used instructional and egocentric video datasets that provide dense and accurate temporal action boundary annotations. The reliance on this format is crucial for deriving objective ground-truth answers: $$(action, t_s, t_e).$$

% YouCook2: long instructional videos annotated with fine-grained cooking steps. 

% COIN: structured instructional tasks with consistent multi-step action boundaries.

% EgoExo4D: synchronized first- and third-person videos with dense, segment-level temporal annotations.

% {\color{red}
% For YouCook2 and COIN, we use the official training splits to construct our training set and the corresponding evaluation splits for building our evaluation set. For EgoExo4D, we use the ``keystep'' annotations and follow its official training and evaluation splits for dataset construction.}

\subsubsection{Source Datasets with Temporal Annotations}
We selected three widely-used instructional and egocentric video datasets that provide dense and accurate temporal action boundary annotations. This high-quality temporal labeling is crucial for deriving objective ground-truth answers in the form of $(action, t_s, t_e)$. 

The composition of our dataset is as follows:
\begin{itemize}
    \item \textbf{YouCook2}: Long instructional videos of cooking steps, contributing approximately 34\% of our samples.
    \item \textbf{COIN}: Structured multi-step tasks, accounting for 40\% of the dataset.
    \item \textbf{EgoExo4D}: Fine-grained "keystep" annotations from egocentric views, making up the remaining 26\%.
\end{itemize}

For YouCook2 and COIN, we utilize the official training and evaluation splits. For EgoExo4D, we follow the standard keystep annotation protocols for split construction. To ensure reproducibility and eliminate LLM-induced hallucinations, the ground-truth CoT for these samples is deterministically synthesized from the human-annotated action sequences using predefined templates, rather than being generated by a large model.

% \subsubsection{Temporal Reasoning Templates}

% To cover a broad range of temporal reasoning phenomena, we design 15 template families, as summarized in Table~\ref{tab:eventflow_templates}. Each template represents a distinct type of temporal relation and contains placeholders (e.g., \{A\}, \{B\}, or \{k\}) that are later instantiated using action labels extracted from the video annotations.

% {\color{red}
% The templates are organized into two hierarchical categories: \textit{action-level} and \textit{video-level} temporal reasoning.

% \textbf{Action-level temporal reasoning} focuses on localized temporal relations around a specific action or short sequence of actions. These questions can typically be answered by reasoning over a limited temporal window rather than the entire video. For example, the template ``What action happened immediately before \{A\}?'' requires identifying the action that directly precedes a given action.

% \textbf{Video-level temporal reasoning}, in contrast, requires global understanding of the entire video or long action sequences. These questions involve reasoning over multiple events and their chronological relationships. For instance, the template ``Arrange these actions in the correct order: \{A, B, C\}.'' requires the model to infer the correct temporal ordering of several actions across the video.

% This hierarchical design enables EventFlowQA to evaluate both \emph{local temporal reasoning} and \emph{global sequence understanding}, providing a comprehensive benchmark for temporal reasoning in video-language models.
% }

\subsubsection{Temporal Reasoning Templates}
To cover a broad range of temporal reasoning phenomena, we design 15 template families, as summarized in Table~\ref{tab:eventflow_templates}. Each template represents a distinct type of temporal relation and contains placeholders (e.g., \{A\}, \{B\}, or \{k\}) that are later instantiated using action labels extracted from the video annotations.

The templates are organized into a hierarchical taxonomy to evaluate two core capabilities:

\textbf{Action-level temporal reasoning} (\textit{$\sim$39.1\% of EventFlowQA}): This category focuses on localized temporal relations around specific actions. These tasks require the model to identify immediate dependencies within a limited temporal window. For example, the \textit{Pre-action} and \textit{Post-action} templates (e.g., ``What happened immediately before \{A\}?'') test the model's precision in boundary localization.

\textbf{Video-level temporal reasoning} (\textit{$\sim$60.9\% of EventFlowQA}): In contrast, these tasks require a global understanding of the entire video or long sequences. This includes \textit{Action Ordering} (e.g., ``Arrange \{A, B, C\} in order''), \textit{Action Counting}, and \textit{Long-term Dependency} reasoning. These questions are specifically designed to challenge the model's ability to maintain a coherent \textit{Summary} and \textit{Think} process without drift over extended horizons.

As shown in Table~\ref{tab:eventflow_templates}, this hierarchical design spans across diverse reasoning types including Sequence, Duration, and Frequency. By mapping human-annotated action boundaries to these templates, we ensure that the resulting ground-truth CoT paths are both factually grounded and structurally rigorous, providing a robust signal for our single-stage RL training.

\begin{table*}[t]
\centering
\caption{Overview of the temporal reasoning templates used in the EventFlow dataset, grouped into action-level and video-level categories.}
\small
\begin{tabular}{p{3.2cm} p{6.2cm} p{7.0cm}}
\toprule
\textbf{Category} & \textbf{Description} & \textbf{Example Template} \\ 
\midrule

\rowcolor{gray!15}
\multicolumn{3}{l}{\textbf{Action-Level Temporal Reasoning}} \\

Action Order Reasoning &
Identify the action that happens immediately before or after a given action. &
\emph{“What action happened immediately before \{A\}?''} \\

Temporal Causality &
Select the action that causes or results from a given action. &
\emph{“What was the most recent action that led to \{A\}?''} \\

Action Anticipation &
Predict the most likely next action following a given step. &
\emph{“What action will the person perform immediately after \{A\}?''} \\

Sequential Prediction &
Choose the next likely action(s) in an ongoing sequence. &
\emph{“Predict the next \{k\} actions after \{A\}.''} \\

Duration Estimation &
Estimate how long an action takes. &
\emph{“How long does the action \{A\} take?''} \\

Temporal Localization &
Find the exact time interval when an action happens. &
\emph{“When does the action \{A\} occur in the video?''} \\

Temporal QA from Narration &
Choose the action that aligns with a narrated timestamp. &
\emph{“What action is taking place at time ratio \{t\}?''} \\

\midrule
\rowcolor{gray!15}
\multicolumn{3}{l}{\textbf{Video-Level Temporal Reasoning}} \\

Temporal Gap Estimation &
Estimate the time gap between two actions. &
\emph{“How much time passed between \{A\} and \{B\}?''} \\

Temporal Yes/No &
Answer whether one action happened before another. &
\emph{“Did \{A\} happen before \{B\}?''} \\

Temporal Comparison &
Choose which action occurred earlier or later. &
\emph{“Which action occurred first?''} \\

Duration Comparison &
Choose the action that took more or less time. &
\emph{“Which action took the longest?''} \\

In-between Action &
Identify actions that happened between two given actions. &
\emph{“Which actions occurred between \{A\} and \{B\}?''} \\

Transcription &
Identify all actions in correct order. &
\emph{“Which option best represents the sequence of actions in the video?''} \\

Action Order Reasoning (Extreme) &
Select the first or last action from a sequence. &
\emph{“Which action happened first in the video?''} \\

Action Order Sorting &
Sort multiple actions in the correct chronological order. &
\emph{“Arrange these actions in the correct order: \{A,B,C\}.''} \\

Order Consistency &
Identify which option maintains correct chronological sequence. &
\emph{“Which option shows the actions in the correct chronological order?''} \\

\bottomrule
\end{tabular}

\label{tab:eventflow_templates}
\end{table*}

\subsubsection{LLM-Driven Question  Generation}
\paragraph{Template Rewriting using an LLM}
To mitigate the risk of RL models overfitting to simple, repetitive sentence structures, we utilized a state-of-the-art LLM to increase the linguistic diversity of our templates.

\begin{itemize}
    \item \textbf{Process}: Each of the 15 core templates was fed to the LLM with a prompt instructing it to generate multiple paraphrased variations (typically 10-15 per template).
    
    \item  \textbf{Constraint}: The LLM was strictly constrained to maintain the exact temporal logic and the placement of the action placeholders, ensuring that the rewritten query remained logically identical to the core template.
    
    \item \textbf{Result}: This process expanded the template pool from 15 to hundreds of unique question structures, significantly enhancing the natural language robustness of the dataset.

\end{itemize}

\paragraph{Action Grounding and Slot Filling}

The final stage involves generating the executable QA pairs by merging the rewritten templates with the video data and deriving the ground-truth answer.
\begin{itemize}
    \item \textbf{Selection and Insertion}: A randomly selected rewritten template is paired with relevant action segments extracted from the source video datasets. The action labels (e.g., "chop vegetables," "heat oil") are inserted into the placeholders.
    
    \item  \textbf{Ground Truth Derivation}: The ground-truth answer is programmatically computed solely based on the temporal logic derived from the video's $t_s$ and $t_e$ annotations. For instance, if a query is of the "Precedes" type, the answer is determined by verifying the condition $t_{e(A)} < t_{s(B)}$ against all possible actions $A$ within the context. This rigorous, data-driven approach ensures the absolute correctness of every generated answer.
\end{itemize}

\subsection{Output Format and Examples}

\begin{figure*}[t]
    \centering
    \includegraphics[width=0.8\linewidth,]{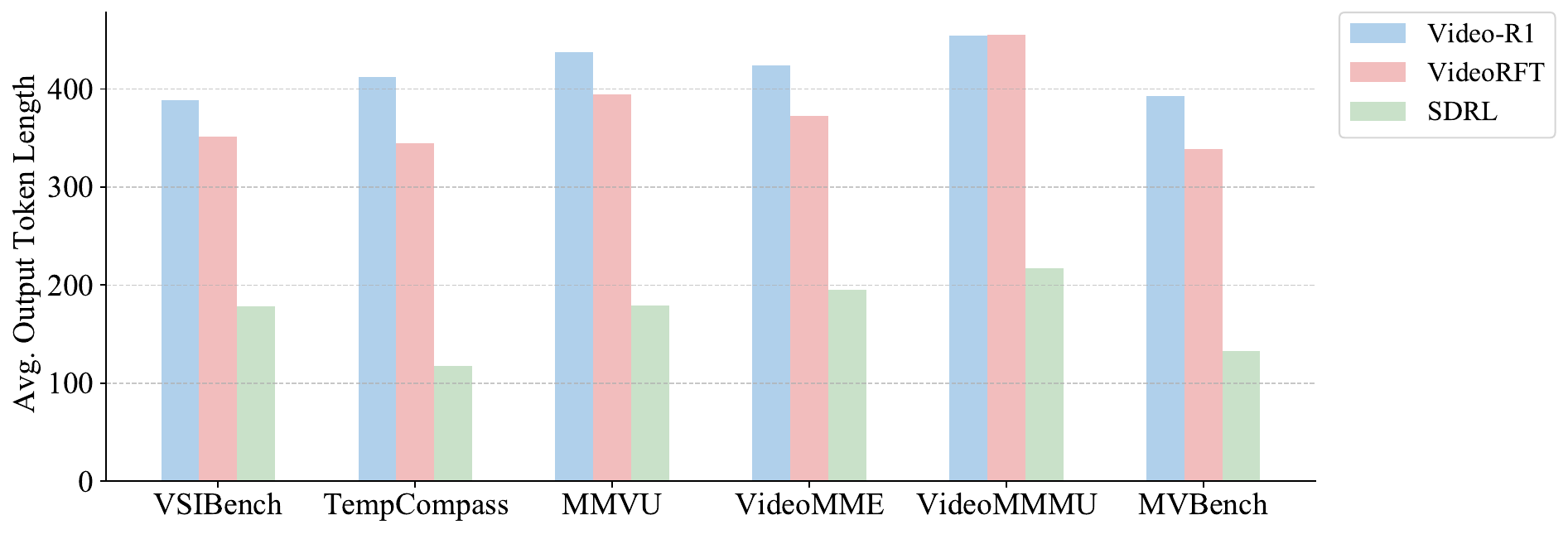}
    \caption{Comparison of the average output token length across six video benchmarks.
We observe that Video-R1 consistently produces longer reasoning sequences than VideoRFT and SDRL, while SDRL yields the most concise outputs across all datasets.}
    \label{fig:avg}
    \vspace{-1mm}
\end{figure*}

As shown in Listing~\ref{lst:json_example}, 
a sample instance of EventFlow  includes the following key information required for training and evaluation: 
% \begin{item}
%     \item video\_id: Unique identifier of the source video.
    
%     \item segment\_id: The specific temporal span used as context.
    
%     \item question: The instantiated, natural language question.
    
%     \item answer: The programmatically derived ground-truth action label, time value, or Boolean response.
    
%     \item query\_type: The core temporal logic category (e.g., Sequential, Durational).
    
%     \item context\_actions: A list of the $$(action, t_s, t_e)$$ triplets used to derive the answer.
% \end{item}

\begin{itemize}
    \item \textbf{Video ID}: Unique identifier of the source video.

    \item \textbf{Video interval}: The specific temporal span used as context.

    \item \textbf{Question}: The instantiated, natural language question.

    \item \textbf{Answer}: The programmatically derived ground-truth action label, time value, or Boolean response.

    \item \textbf{Query type}: The core temporal logic category (e.g., Sequential, Durational).
\end{itemize}

% \begin{lstlisting}[language=json, caption={Example JSON entry from our constructed dataset.}, label={lst:json_example}]
\begin{lstlisting}[caption={Example JSON entry from our constructed dataset.}, label={lst:json_example}]
{
  "tag": "action level",
  "type": "temporal_neighbor",
  "template_type": "MCQ",
  "answer_type": "choice",
  "category": "Action Order Reasoning",
  "template": "What action happened immediately before {} in the video?",
  "new_template": "What activity took place directly prior to {} in the video?",
  "question_args": [
    "'wipe screen again'"
  ],
  "choices": "A. paste protector on the screen\nB. remove the label\nC. place label\nD. wipe screen",
  "final_answer": "B",
  "answer_text": "remove the label",
  "video_interval": [32.39, 134.19],
  "act_idx": 4
}
\end{lstlisting}

\subsection{Data Distribution Analysis}
\begin{figure}[!h]
    \centering
    \includegraphics[width=1\linewidth]{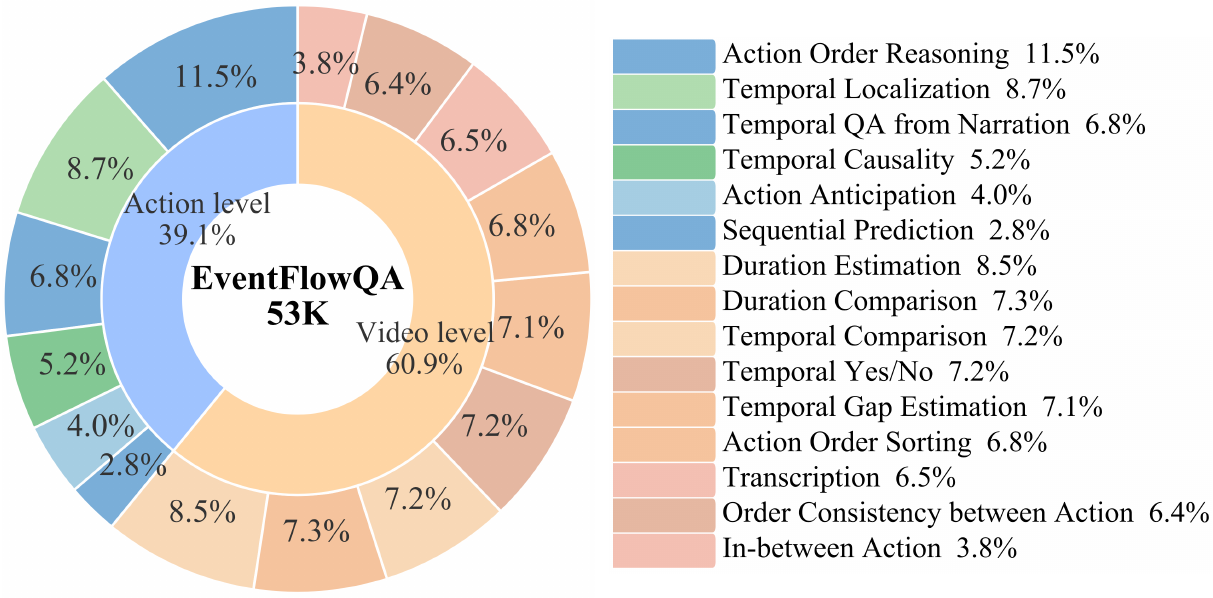}
    \caption{Data distribution of the EventFlowQA-53K dataset.
The dataset consists of two hierarchical levels: Action-level questions and Video-level questions. The outer ring shows the detailed proportions of each temporal reasoning category, including action order, duration, comparison, causality, anticipation, and others.}
    \label{fig:data_dis}
\end{figure}

To provide a granular understanding of EventFlowQA, we analyze the distribution of question types, as illustrated in Fig.~\ref{fig:data_dis}. The dataset comprises a 50K training split and a 3K evaluation split. It is structured into two hierarchical levels of temporal reasoning: \textit{action-level} and \textit{video-level}. Action-level questions account for 39.1\% of the dataset, focusing on localized reasoning over specific actions or short temporal segments. In contrast, video-level questions constitute the majority (60.9\%), requiring a holistic understanding of long-range temporal contexts and inter-event relationships.

The fine-grained distribution across 15 temporal reasoning categories is depicted in the outer ring of Fig.~\ref{fig:data_dis}. These categories span a diverse range of skills, including action ordering, duration estimation, and causal reasoning. Notably, \textit{Action Order Reasoning} (11.5\%), \textit{Temporal Localization} (8.7\%), and \textit{Duration Estimation} (8.5\%) are the most prominent categories, reflecting the dataset's emphasis on capturing the underlying temporal structure of videos.

Furthermore, challenging scenarios such as \textit{Sequential Prediction}, \textit{Action Anticipation}, and \textit{In-between Action} are included to evaluate a model's ability to infer implicit temporal relations. This multi-faceted distribution ensures that EventFlowQA serves as a comprehensive and demanding benchmark for assessing the temporal reasoning capabilities of modern video-language models.

\section{Prompt For Training and Inference}
Our prompt used for both training and inference is shown in Table~\ref{tab:prompt}. It introduces a summary tag to encourage the model to summarize the video before generating its reasoning and final answer.

\begin{table}[h]
\centering
\caption{Prompt for training and inference.}
\begin{tabularx}{\columnwidth}{p{2.5cm}X}\toprule
        % \rowcolor{gray!15} 
        \multicolumn{2}{l}{\textbf{Prompt Template}} \\ \midrule

        \multicolumn{2}{p{\columnwidth}}{ \{Question\} \{Type of Template\}
        Given the video clip and the question, first summarize the sequence of actions observed in the video using \textless summary\textgreater  \textless /summary\textgreater tags. 
        Then, use that summary to guide your reasoning inside \textless think\textgreater \textless/think\textgreater tags. 
        Finally, state the answer clearly in \textless answer\textgreater \textless/answer\textgreater tags.}\\ \midrule\midrule
        % \rowcolor{gray!15} 
        \multicolumn{2}{l}{\textbf{Type of Template}}\\ \midrule
        multiple choice&  Please provide only the single option letter (e.g., A, B, C, D, etc.).\\ \midrule
        numerical & Please provide the numerical value (e.g., 42 or 3.14).\\ \midrule
        OCR & Please transcribe text from the image/video clearly and provide your text answer.\\ \midrule
        free-form & Please provide your text answer.\\ \midrule
        regression & Please provide the numerical value (e.g., 42 or 3.14).\\
\bottomrule
\label{tab:prompt}
\end{tabularx}
\end{table}

\section{Different Tags for Summary}
\begin{table*}[t]
\centering
\caption{Effect of different prompt tags on model behavior.
For each tag, the left column shows the question prompt and the right column shows the model output. Accuracy values reflect the QA performance under each tag. The examples show how prompt structure shapes the granularity, ordering, and coherence of the generated summaries.}
\begin{tabularx}{\textwidth}{>{\raggedright\arraybackslash}p{7.8cm} X}
\toprule

\multicolumn{2}{c}{
    \includegraphics[width=0.9\textwidth]{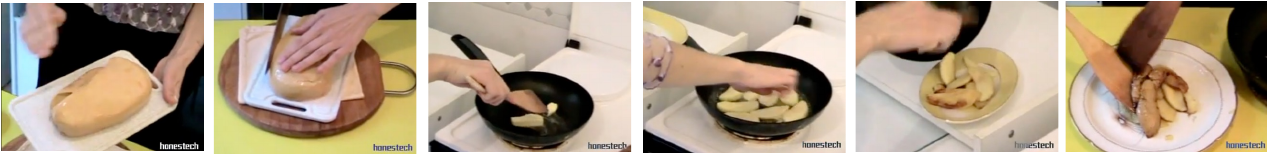}
} \\
\midrule

\textbf{Question Prompt} (tag, accuracy) & \textbf{Model Output} \\ \midrule

 \colorbox{gray!20}{\texttt{\textbf{action}}} (26.45\%)\\
 Given a video clip, \textbf{first list the sequence of actions observed in the video in order in \textless action\textgreater \textless /action\textgreater tags.}
 Then, answer the question with output the thinking process in \textless think\textgreater \textless /think \textgreater tags, followed by the final answer in \textless answer\textgreater \textless /answer \textgreater tags.

     &
      \textbf{\textless action\textgreater }1. put 2 tsp of olive oil add cut bread dices and fry 2. tear fresh lettuce leaves and add them with seasoned kale leaves 3. cook bacon in a pan until crispy 4. add roasted bread crumbs crispy bacon slices of parmesan cheese  shaved  to seasoned greens and pour dressing blend on top\textbf{\textless /action\textgreater}
 \\ \midrule

\colorbox{gray!20}{\texttt{\textbf{caption}}} (26.95\%) \\
 Given a video clip, \textbf{first describe the sequence of actions observed in the video in the exact order they occur using \textless caption\textgreater \textless caption\textgreater tags.}  Then, answer the question with output the thinking process in \textless think\textgreater \textless /think \textgreater tags, followed by the final answer in \textless answer\textgreater \textless /answer \textgreater tags.

&
      \textbf{\textless caption\textgreater} A woman is standing in her kitchen. She picks up a piece of bread and places it on a cutting board. She then cuts the bread into slices. Next, she puts some butter in a frying pan and cooks the foie gras. After that, she cooks the apple slices in the same pan. Finally, she places the apple slices and the foie gras on a plate.\textbf{\textless /caption \textgreater}

 \\ \midrule

\colorbox{gray!20}{\texttt{\textbf{event}}} (16.12\%) \\
\multirow{6}{=}{%
\begin{minipage}[t]{\linewidth}
 Given a video clip, \textbf{first describe the sequence of events observed in the video in the exact order they occur using \textless event\textgreater \textless event\textgreater tags.}  Then, answer the question with output the thinking process in \textless think\textgreater \textless /think \textgreater tags, followed by the final answer in \textless answer\textgreater \textless /answer \textgreater tags.
 \end{minipage}
}
 
&
            \textbf{\textless event\textgreater} cut the foie gras into slices \textbf{\textless /event\textgreater}  \\
            & \textbf{\textless event\textgreater} pour the juice onto the foie gras\textbf{\textless /event\textgreater}  \\
            &\textbf{\textless event\textgreater} place the apples and the foie gras on a plate\textbf{\textless /event\textgreater}  \\
            &\textbf{\textless event\textgreater} sprinkle salt and pepper on top\textbf{\textless /event\textgreater} 
            \\
 \\ \midrule

\colorbox{gray!20}{\texttt{\textbf{segment}}}  (23.53\%)\\
\multirow{5}{=}{%
\begin{minipage}[t]{\linewidth}
Given a video clip, \textbf{first summarize the sequence of actions observed in the video using \textless segment\textgreater \textless /segment\textgreater tags.}
Then, output the thinking process in \textless think\textgreater \textless /think\textgreater tags, 
followed by the final answer in \textless answer\textgreater \textless /answer\textgreater tags.
\end{minipage}
}
&
      \textbf{\textless segment\textgreater}0:12.3 - 0:14.6\textbf{\textless /segment\textgreater}\\
     &\textbf{\textless segment\textgreater}0:15.0 - 0:17.8\textbf{\textless /segment\textgreater}\\
      &\textbf{\textless segment\textgreater}0:19.0 - 0:20.2\textbf{\textless /segment\textgreater}\\
      &\textbf{\textless segment\textgreater}0:20.6 - 0:21.7\textbf{\textless /segment\textgreater}\\
      &...... \\ \midrule

\colorbox{gray!20}{\texttt{\textbf{summary}}} (30.96\%)  \\
 Given a video clip, \textbf{first summarize the sequence of actions observed in the video using \textless summary\textgreater \textless /summary\textgreater tags.}
 Then, answer the question with output the thinking process in \textless think\textgreater \textless /think\textgreater tags,  followed by the final answer in \textless answer\textgreater \textless /answer\textgreater tags.
&
      \textbf{\textless summary\textgreater}      
      A woman is preparing a dish in her kitchen. She starts by cutting a piece of bread and placing it on a plate. Then, she cuts an apple and places it on the same plate. Next, she cooks the bread in a pan with butter. After that, she cooks the apple slices in another pan. Finally, she places the cooked bread and apple slices on a plate.
 \textbf{\textless /summary\textgreater}
\\
\bottomrule
\end{tabularx}
\label{tab:tags}
\end{table*}

To encourage the model to produce temporally ordered summaries, we explore a series of tags that explicitly guide the model’s initial summarization behavior, shown in Table~\ref{tab:tags}. Our objective is to find a tag that naturally induces the model to output the sequence of actions in the correct chronological order, even before applying any reinforcement learning. In other words, a suitable tag should enable the model to produce summaries with high initial accuracy and clear temporal structure.

We experiment with multiple instruction tags, each designed to prompt the model toward a different style of temporal decomposition, for example, requesting the model to “list the sequences of \textbf{actions},” “describe the sequence of \textbf{events}” or “\textbf{summarize} the sequence of actions.” Although these tags all aim at temporal organization, they lead to substantially different model behaviors.

For each tag, we compute the initial accuracy. This quantitative comparison on Table~\ref{tab:tags} allows us to assess which tag most effectively supports temporal grounding prior to RL optimization. To further understand the qualitative differences introduced by each tag, we visualize representative summaries generated under different tag settings. 
Based on this combined quantitative and qualitative analysis, we select the tag that achieves the highest initial accuracy. This tag subsequently serves as the default instruction during structured summary generation in our framework.

\section{Efficiency Analysis}

% \begin{figure*}[t]
%     \centering
%     % \includegraphics[width=\linewidth,height=.55\linewidth]
%     \includegraphics[width=0.7\linewidth,]{tokens_std.pdf}
%     \caption{Comparison of the output token length variance across six video benchmarks.
% VideoRFT shows the highest variance, indicating unstable reasoning lengths across samples, whereas SDRL produces the most stable and consistent output lengths.}
%     \label{fig:std}
%     \vspace{-1mm}
% \end{figure*}

\textbf{Analysis of Output Token Length.} Our method, SDRL, driven by auxiliary summaries and optimized via Reinforcement Learning (RL), exhibits a clear and beneficial characteristic in its generated output (CoT/Summary): a \textbf{significantly shorter average token length} compared to baseline MLLMs (Video-R1 and VideoRFT), as demonstrated in Figure~\ref{fig:avg}. This statistical finding is crucial for interpreting the model's behavior and the effectiveness of our optimization strategy in complex video understanding tasks.

\begin{itemize}
    \item \textbf{Efficiency and Conciseness} 
    As shown in Figure~\ref{fig:avg}, the average output length of SDRL is consistently the shortest across all six video benchmarks. This strong indication of conciseness demonstrates the model’s overall \textbf{high efficiency} in generating explanations.
    
\item \textbf{Successful Consistency in Summary}
    The RL framework, guided by the summary, effectively regularizes the model's output generation process. This optimization encourages the MLLM to prune redundant or repetitive reasoning steps, focusing only on the minimum essential information required for accurate prediction.
    
    \item \textbf{High-Quality Outputs}
    The reduced token count is not merely compression, coupled with our superior performance, this short length confirms a successful optimization towards \textbf{high-quality outputs}. This characteristic directly translates to practical advantages, including reduced inference latency and lower computational overhead during deployment.
\end{itemize}

In summary, the substantial reduction in average output length confirms that SDRL learns a highly effective and concise reasoning path, demonstrating that our RL optimization successfully yields a model that is both powerful and efficient.

% {\color{red}
% \section{Additional Analysis of Self-Supervised CVK Stability}

% A potential concern of the self-supervised CVK objective is whether enforcing consistency among sampled summaries could lead to model collapse or reinforce hallucinated summaries. In this section, we provide additional empirical analysis showing that improved consistency does not lead to such failure modes.

% First, as shown in Table~4 of the main paper, self-supervised CVK improves summary factual alignment, achieving +3.37\% BLEU and +9.43\% sBERT compared to the baseline without CVK. These improvements indicate stronger semantic agreement between generated summaries and ground-truth summaries.

% Second, Figure~6 in the main paper provides qualitative examples demonstrating that SDRL produces summaries that more accurately reflect the temporal structure of the video compared to prior RL-based approaches.

% Finally, we track the summary quality score during training, measured as $(\text{BLEU}+\text{sBERT})/2$. As shown in Fig.~\ref{fig:summary_quality_curve}, the score increases steadily throughout training without any evidence of collapse or degeneration, suggesting that the self-supervised consistency objective stabilizes rather than harms the reasoning process.}

\section{Stability and Factual Grounding of CVK}

A key concern regarding self-supervised consistency is whether enforcing agreement among sampled summaries might lead to ``mode collapse'' or the reinforcement of hallucinations. We provide empirical evidence to show that our CVK objective remains stable and factually grounded throughout training.

\textbf{Mechanism of Stability.} As defined in Eq.~\ref{eq:anchor} in the main paper, the consistency anchor is derived exclusively from the group members that yield the \textit{correct} final answer. This filtering mechanism ensures that the model aligns towards reasoning paths that are not only consistent but also functionally effective, preventing the mutual reinforcement of erroneous summaries during early training stages.

\textbf{Quantitative Factuality.} As shown in Table~\ref{tab:cvk} in the main paper, the self-supervised CVK significantly enhances summary quality, achieving a +3.37\% improvement in BLEU and +9.43\% in sBERT compared to the baseline. These gains indicate that enforcing consistency among ``correct'' paths naturally encourages the model to capture more accurate semantic and temporal information from the video.

\begin{figure}[t]
\centering
\includegraphics[width=0.85\linewidth]{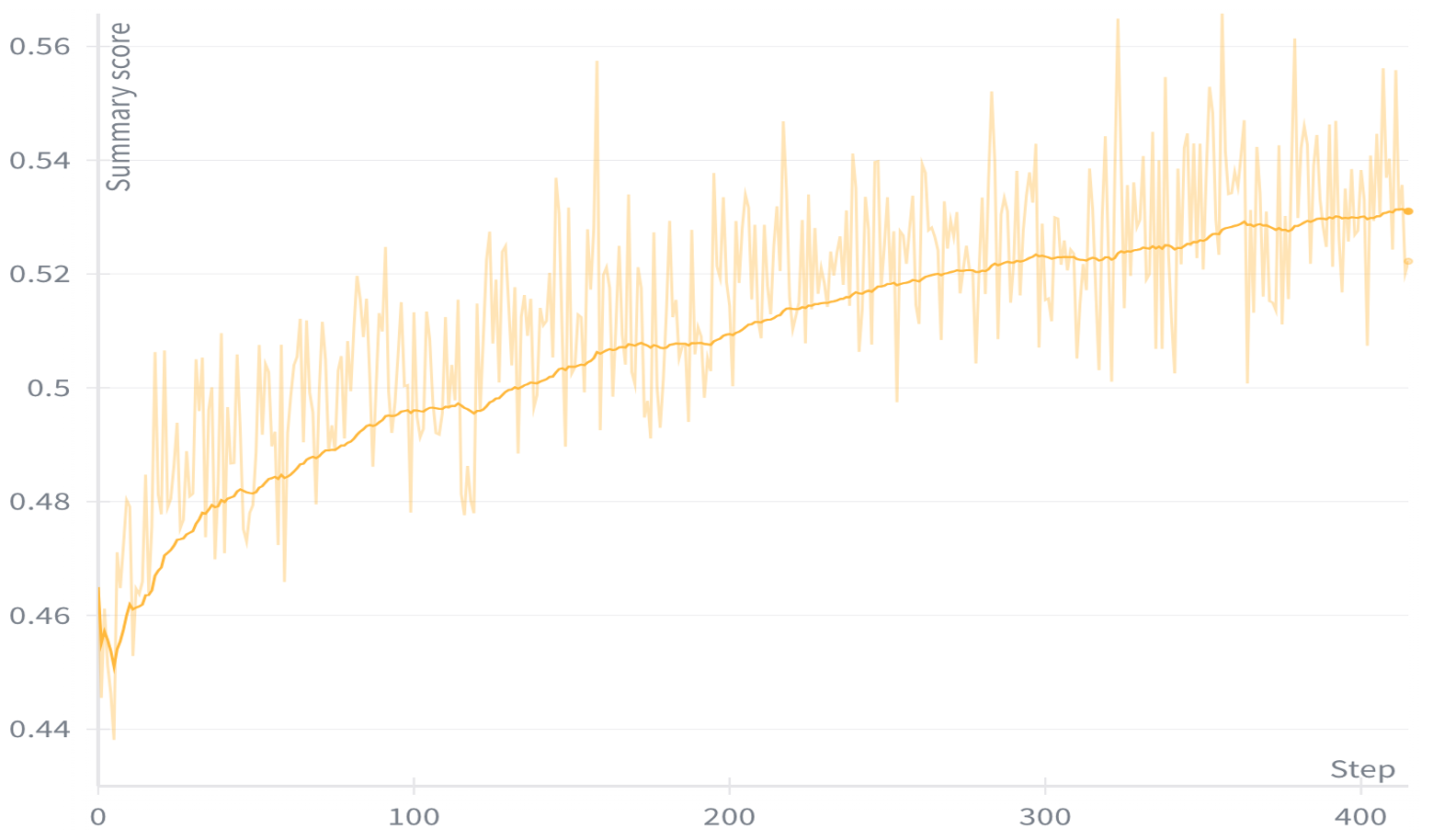}
\caption{Summary quality during training measured by $(\text{BLEU}+\text{sBERT})/2$. The curve shows steady improvement without collapse.}
\label{fig:summary_quality_curve}
\end{figure}

\textbf{Training Dynamics.} We monitor the summary quality score, $(\text{BLEU}+\text{sBERT})/2$, across the training process. As illustrated in Fig.~\ref{fig:summary_quality_curve}, the quality score improves monotonically and stabilizes as the model converges. There is no evidence of performance degradation or collapse, confirming that CVK effectively regularizes the reasoning space without sacrificing factual integrity. Qualitative examples in Fig.~6 further demonstrate that SDRL produces more temporally-precise summaries than previous RL-based methods.

\section{More Visualization}
% \subsection{On EventFlow}

% \subsection{On Video-r1-260k}

\begin{figure*}[t]
    \centering
    \includegraphics[width=\linewidth,]{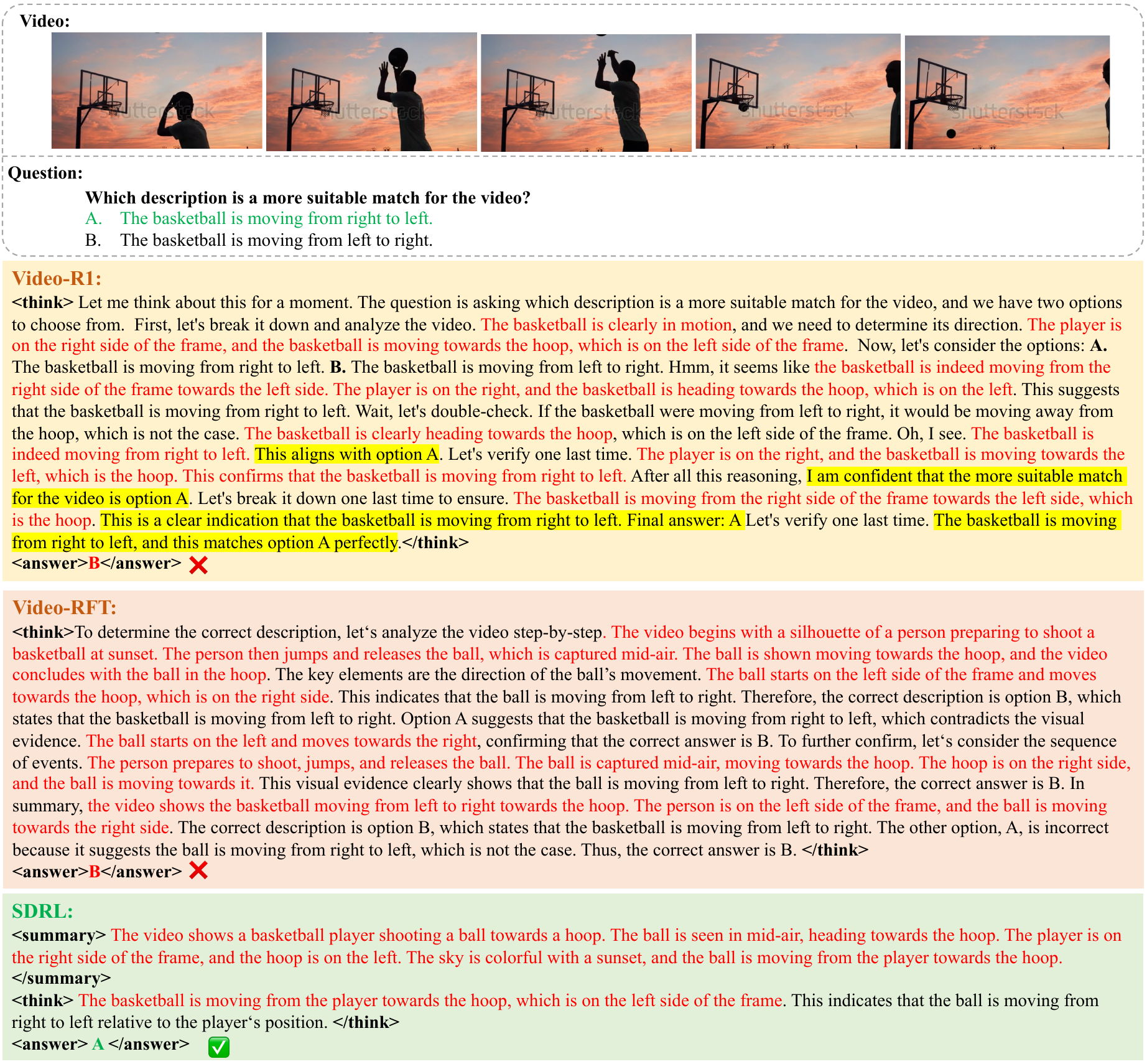}
    \caption{An example on Temcompass dataset. Video-R1 and Video-RFT generate lower information density descriptions (\textcolor{red}{All red part is the video description}). Video-R1's description is iterative, which may contribute to a longer reasoning process. This extended path may be associated with the observed instability in the inference chain \colorbox{yellow}{(yellow highlighting)}, where intermediate thoughts contradict the final outcome, resulting in the selection of the incorrect answer. Video-RFT's context is more dispersed, which potentially leads to an initial misrepresentation of the visual facts.
SDRL produces a concise, highly focused summary by directly extracting the necessary visual facts ("ball on the right, hoop on the left"). This concentrated context supports a shorter and more direct reasoning path.}
    \label{fig:vis_mmvu54}
    \vspace{-1mm}
\end{figure*}

\begin{figure*}[t]
    \centering
    \includegraphics[width=\linewidth,]{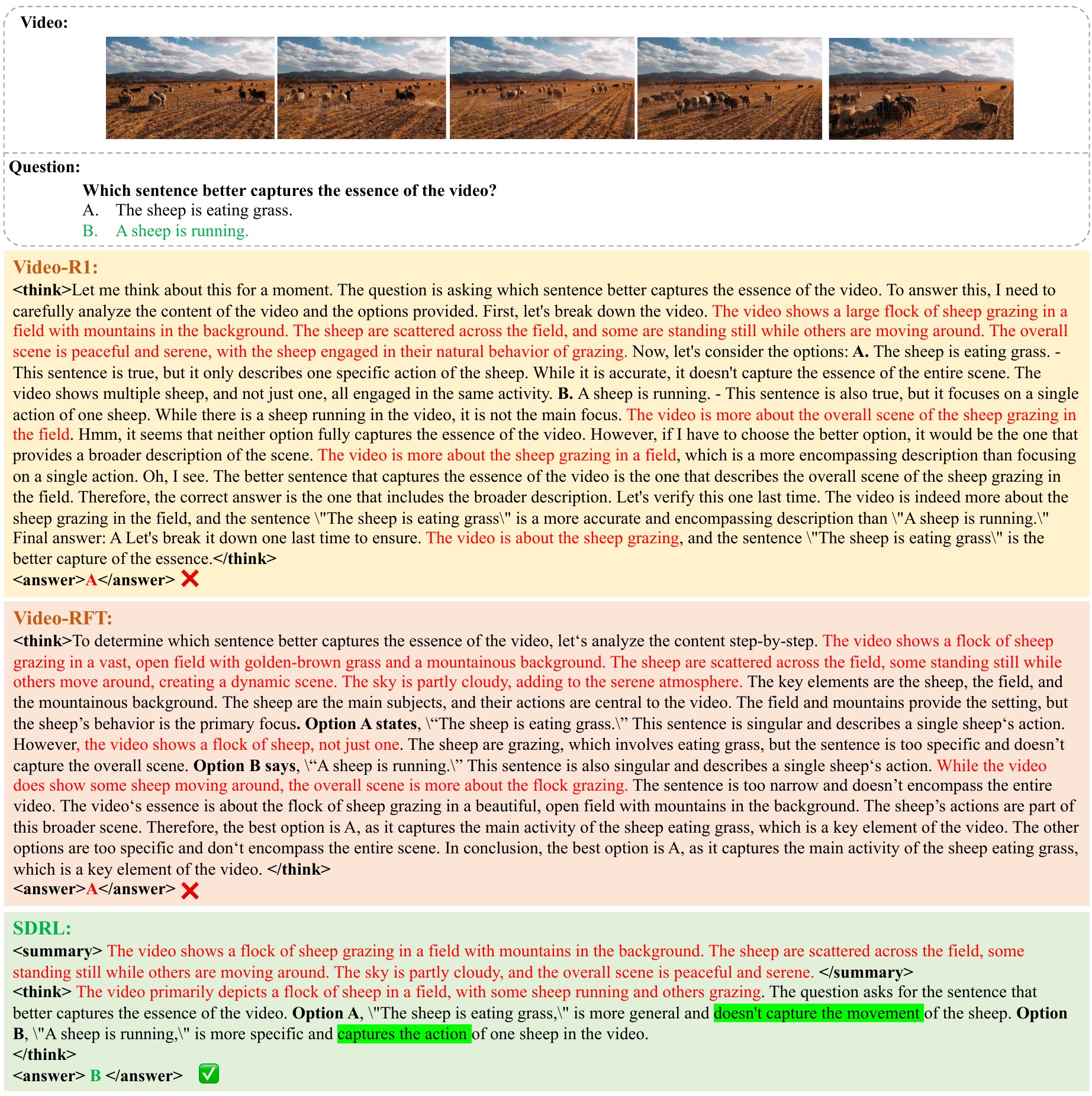}
    \caption{An example on Temcompass dataset. Although all three methods describe roughly the same scene (\textcolor{red}{descriptions in red}), their narrative focus differs significantly. SDRL produces a concise and coherent summary that preserves global context, enabling the model to capture the overall movement pattern of the sheep (\colorbox{green}{green highlighting}) and correctly identify the key action in the video. In contrast, Video-R1 and Video-RFT provide fragmented or overly broad descriptions that dilute the salient motion cues, causing the reasoning to focus on partial details and miss the essential dynamics of the scene.}
    \label{fig:vis_temcompass}
    \vspace{-1mm}
\end{figure*}

\begin{figure*}[t]
    \centering
    \includegraphics[width=\linewidth,]{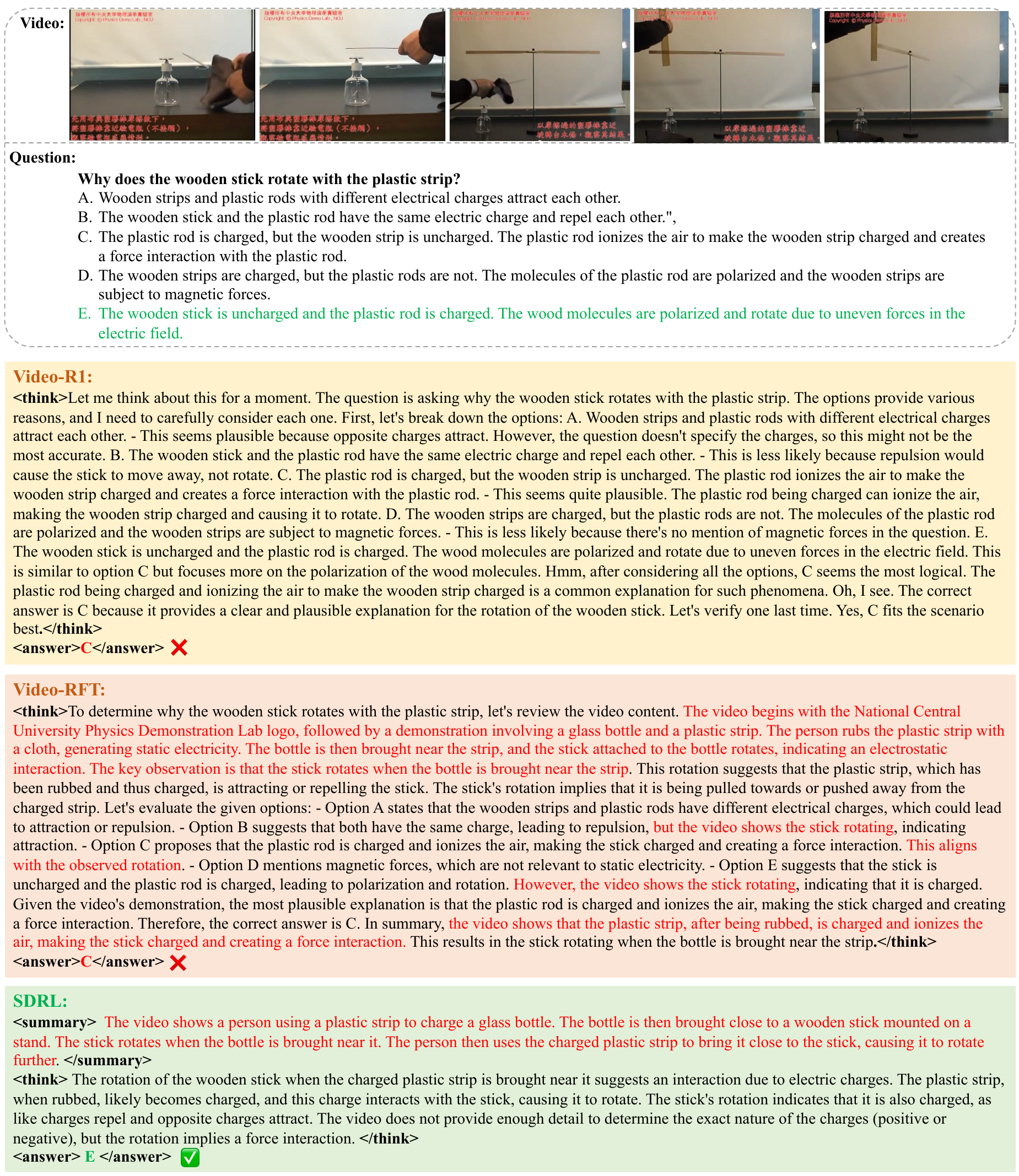}
    \caption{An example on MMVU dataset. Video-R1 provides almost no grounding in the actual video content and instead jumps directly into option analysis, making it highly susceptible to thinking drift. Video-RFT includes an initial video description, but its reasoning process gradually shifts away from the visual evidence, leading to potential drift as well. SDRL maintains tight grounding by consistently referencing key visual cues throughout the reasoning process, thereby preventing drift and enabling more reliable decision-making.}
    \label{fig:vis_mmvu51}
    \vspace{-1mm}
\end{figure*}

\begin{figure*}[t]
    \centering
    \includegraphics[width=\linewidth,]{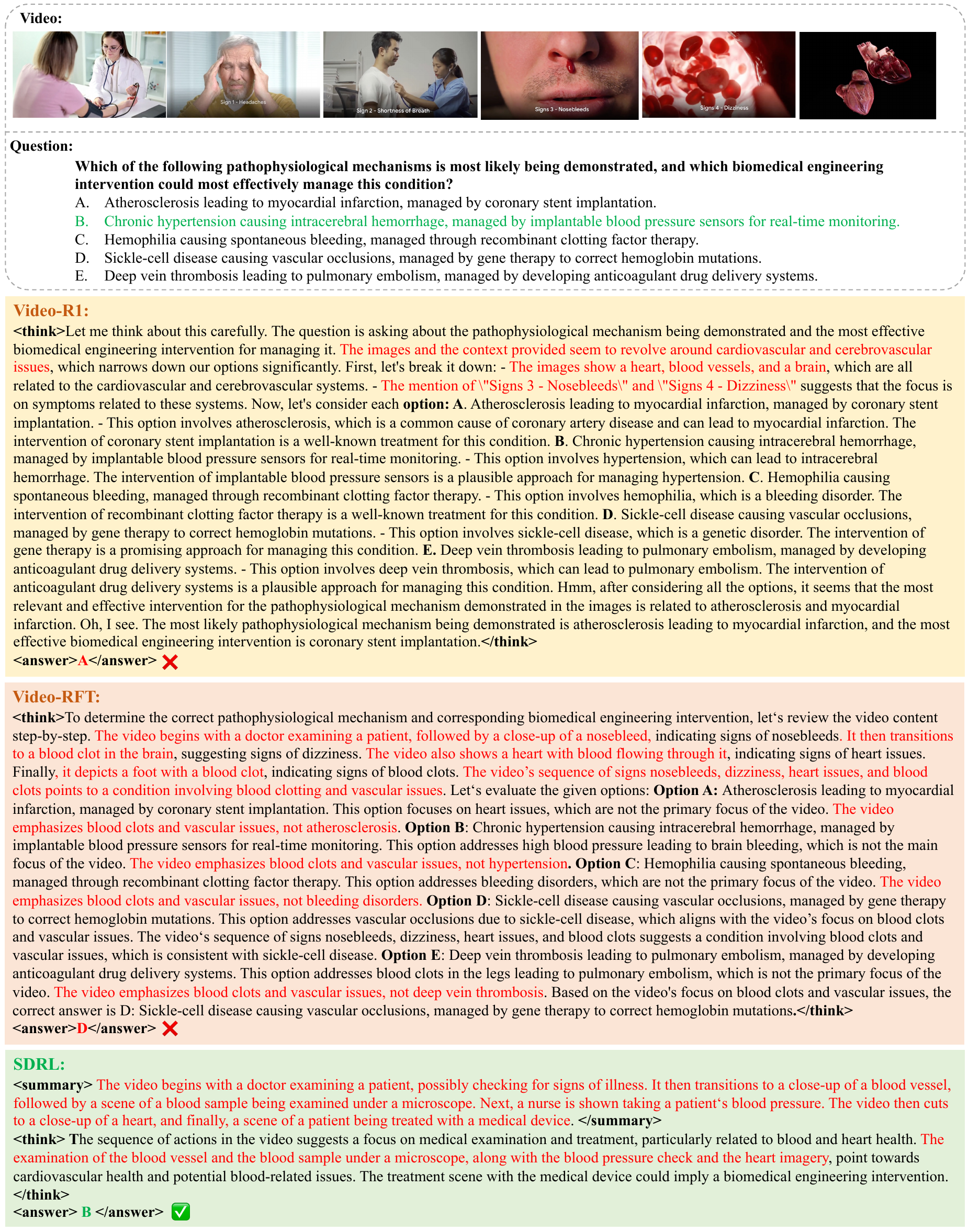}
    \caption{An example on MMVU dataset. Both Video-R1 and VideoRFT devote excessive attention to analyzing the answer options, producing overly long reasoning chains that deviate from the actual visual evidence and easily lead to thinking drift. In contrast, SDRL anchors the reasoning in the video content, maintains a concise and focused explanation, and thus avoids unnecessary detours caused by option-driven overanalysis.}
    \label{fig:vis_mmvu41}
    \vspace{-1mm}
\end{figure*}

We provide qualitative visualizations across several datasets to illustrate the behavior of different reasoning strategies under various video understanding scenarios. Figures \ref{fig:vis_mmvu54}, \ref{fig:vis_temcompass}, \ref{fig:vis_mmvu51}, and ~\ref{fig:vis_mmvu41}
 show examples from different benchmarks. Each visualization includes the input video frames, the corresponding question and answer choices, and the generated reasoning outputs from Video-R1, Video-RFT, and our SDRL. Across diverse reasoning tasks, SDRL consistently produces concise and well-structured summaries that focus on the most salient visual cues necessary for the task. These visualizations highlight how SDRL’s summary-driven reasoning encourages grounded, task-relevant interpretation of the video, providing clearer and more consistent reasoning trajectories compared to the other baselines.

\section{Failure Case}

\begin{figure*}[t]
    \centering
    \includegraphics[width=\linewidth,]{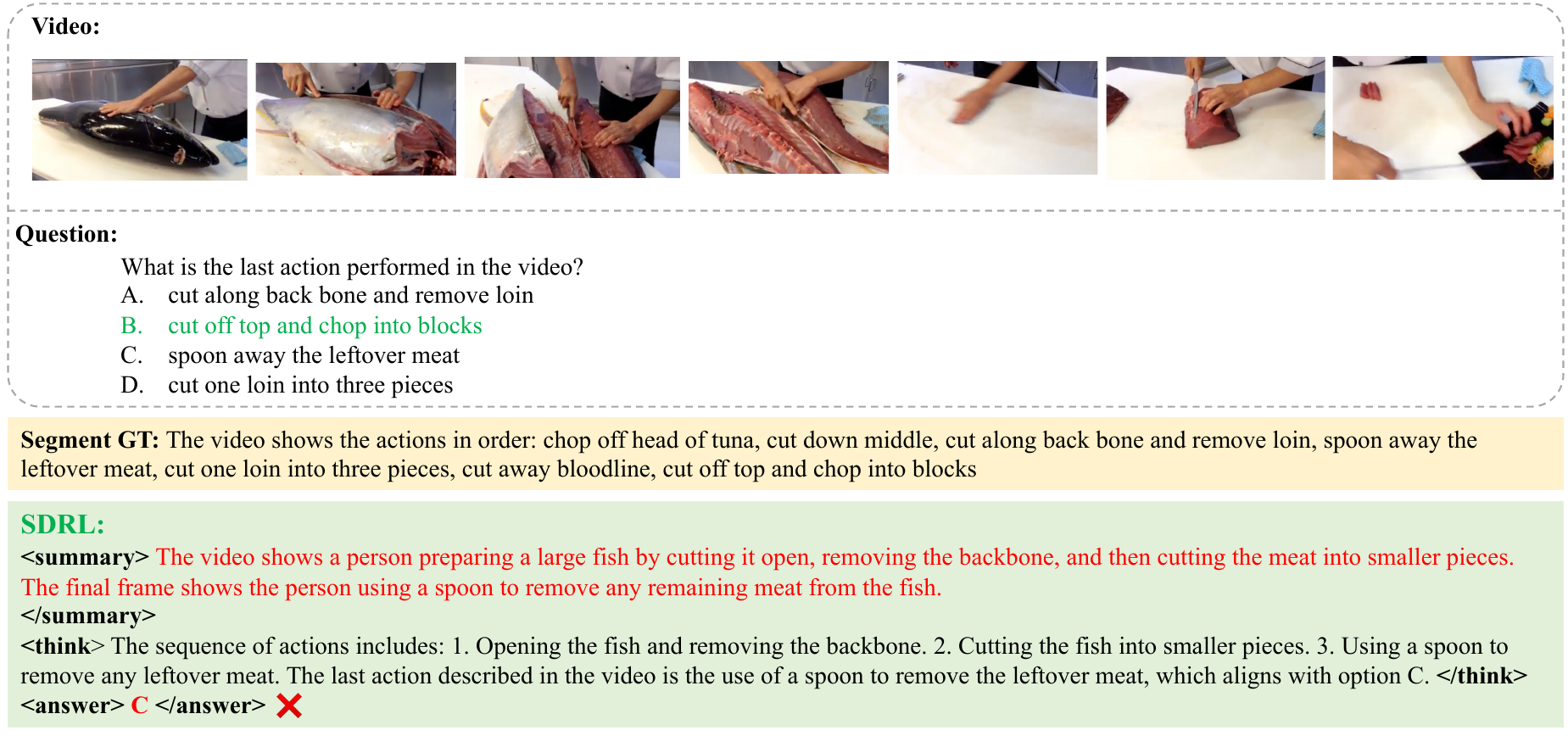}
    \caption{A failure case from EventFlow dataset. Without ground-truth supervision, SDRL may produce incomplete or temporally misaligned summaries, particularly when multiple actions in the video exhibit similar visual patterns.}
    \label{fig:fal_eventflow}
    \vspace{-1mm}
\end{figure*}

To illustrate the limitations of summary-driven reasoning without explicit ground-truth supervision, we analyze a representative failure case shown in Figure~\ref{fig:fal_eventflow}. Although the video contains a long sequence of visually similar cutting actions, the SDRL model generates a summary that omits several intermediate steps and mistakenly focuses on a salient but non-final “spooning” action. This incomplete and temporally misaligned summary propagates into the reasoning process, leading the model to incorrectly select option C, despite the true final action being “cut off top and chop into blocks.” The example highlights a core challenge of unsupervised summary generation: when low-level visual patterns are similar across actions, the model may gravitate toward semantically plausible but temporally incorrect interpretations, ultimately harming downstream question answering.

\begin{figure*}[t]
    \centering
    \includegraphics[width=\linewidth,]{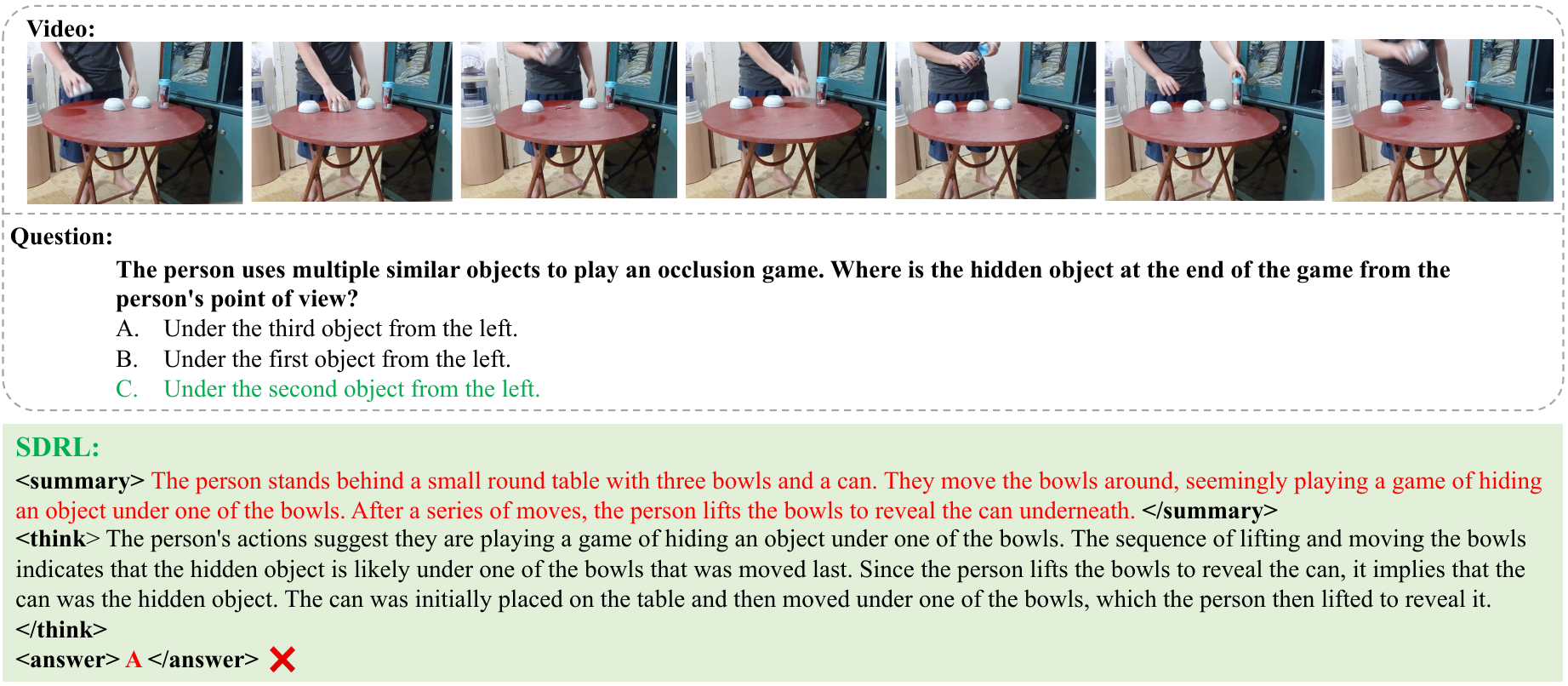}
    \caption{A failure case from MVBench dataset. When actions become highly fine-grained, SDRL often generates coarse, high-level summaries instead of enumerating precise motion steps.}
    \label{fig:fal_mvbench}
    \vspace{-1mm}
\end{figure*}

Moreover, as shown in Figure~\ref{fig:fal_mvbench}, when the video involves fast and fine-grained object manipulations, SDRL tends to produce a coarse, high-level description rather than enumerating each atomic action. In this example of an occlusion game, the model summarizes the sequence as “moving bowls around,” without tracking the precise order of swaps and lifts that determine the final hidden location. This loss of fine-grained temporal detail causes the reasoning chain to rely on a generic interpretation of the game instead of the actual step-by-step motion, ultimately leading SDRL to infer an incorrect hiding position. The case demonstrates that, without explicit supervision on action boundaries, SDRL summaries may abstract away critical micro-actions, resulting in temporally drifting or semantically incomplete reasoning.

\section{Limitations and Future Work}
Despite the effectiveness of our SDRL framework in improving temporal reasoning, it still exhibits certain limitations. First, in scenarios involving long-horizon videos with dense or fine-grained actions, the model may generate relatively coarse summaries that omit micro-level temporal cues. This abstraction bias can lead to temporal drift when tasks demand precise frame-level tracking. Second, while CVK enforces consistency among high-quality reasoning paths, it does not provide a formal guarantee of absolute factual correctness. In cases where answer supervision is sparse, there remains a theoretical risk that the model might converge on consistent but partially hallucinated summaries. Furthermore, our single-stage RL framework relies on the base model’s inherent instruction-following capabilities to parse structured tags; adapting SDRL to significantly weaker models without such priors may require a brief warm-up phase.

Future work will explore incorporating stronger temporal supervision, such as contrastive action boundaries and structured sequence alignment, to encourage finer-grained summary generation. Additionally, we plan to integrate external verification signals or human-in-the-loop evaluation to further anchor the factuality of the reasoning process and investigate multi-step self-refinement strategies to mitigate summary drift in densely interactive video scenarios.

\end{document}